\RequirePackage{fix-cm}
\documentclass[smallcondensed]{svjour3}     % one column (ditto)
\smartqed  % flush right qed marks, e.g. at end of proof
\usepackage{graphicx}
\usepackage[linkcolor=blue]{hyperref}
\usepackage{natbib}
\usepackage{microtype}
\usepackage{bm}
\usepackage{algorithm}
\usepackage[]{algpseudocode}
\usepackage{amsmath}
\usepackage{tikz}
\usetikzlibrary{bayesnet, fit, calc, arrows, decorations.pathmorphing, backgrounds, positioning, matrix}
\usepackage{subcaption}
\usepackage{amsfonts}
\usepackage{booktabs}
\usepackage{multirow}
\usepackage{array}
\usepackage{xcolor}
\usepackage{soul}
\usepackage[title]{appendix}
\usepackage{breqn}
\usepackage{thmtools,thm-restate}
\usepackage[framemethod=tikz]{mdframed}
\definecolor{superlightred}{HTML}{F5F5F5}

\makeatletter
\newcommand*{\indep}{%
  \mathbin{%
    \mathpalette{\@indep}{}%
  }%
}
\newcommand*{\nindep}{%
  \mathbin{%                   % The final symbol is a binary math operator
    \mathpalette{\@indep}{\not}% \mathpalette helps for the adaptation
  }%
}
\newcommand*{\@indep}[2]{%
  \sbox0{$#1\perp\m@th$}%        box 0 contains \perp symbol
  \sbox2{$#1=$}%                 box 2 for the height of =
  \sbox4{$#1\vcenter{}$}%        box 4 for the height of the math axis
  \rlap{\copy0}%                 first \perp
  \dimen@=\dimexpr\ht2-\ht4-.2pt\relax
  \kern\dimen@
  {#2}%
  \kern\dimen@
  \copy0 %                       second \perp
}

\newcommand\addvmargin[1]{
  \node[fit=(current bounding box),inner ysep=#1,inner xsep=0]{};
}

\newtheorem{assumption}{Assumption}

\algnewcommand\algorithmicforeach{\textbf{for each}}
\algdef{S}[FOR]{ForEach}[1]{\algorithmicforeach\ #1\ \algorithmicdo}

\let\oldReturn\Return
\renewcommand{\Return}{\State\oldReturn}

\newcommand{\tikzmark}[1]{\tikz[overlay,remember picture,baseline] \node [anchor=base] (#1) {};}%

\def\drawredbox{%
\begin{tikzpicture}[remember picture, overlay]
\draw[thick, dashed, red, rounded corners] (50pt, 220.5pt) rectangle (345pt, 148pt);
\end{tikzpicture}%
}
\def\HiLi{\leavevmode\rlap{\hbox to 0.015\hsize{\color{yellow!50}\leaders\hrule height .8\baselineskip depth .4ex\hfill}}}

\begin{document}

\title{Greedy structure learning from data that contain systematic missing values%\thanks{Grants or other notes
}


\author{Yang Liu         \and
        Anthony C. Constantinou %etc.
}


\institute{Yang Liu\and Anthony C. Constantinou\at Bayesian Artificial Intelligence research lab, School of Electronic Engineering and Computer Science, Queen Mary University of London, London, UK\\\email{\{yangliu, a.constantinou\}}@qmul.ac.uk}
%\institute{Yang Liu\\\email{yangliu@qmul.ac.uk}\\
%Anthony C. Constantinou\\\email{a.constantinou@qmul.ac.uk}\\
%Bayesian Artificial Intelligence research lab, School of Electronic Engineering and Computer Science, Queen %Mary University of London, London, UK
%}

\date{Received: date / Accepted: date}

\maketitle

\begin{abstract}
Learning from data that contain missing values represents a common phenomenon in many domains. Relatively few Bayesian Network structure learning algorithms account for missing data, and those that do tend to rely on standard approaches that assume missing data are missing at random, such as the Expectation-Maximisation algorithm. Because missing data are often systematic, there is a need for more pragmatic methods that can effectively deal with data sets containing missing values not missing at random. The absence of approaches that deal with systematic missing data impedes the application of BN structure learning methods to real-world problems where missingness are not random. This paper describes three variants of greedy search structure learning that utilise pairwise deletion and inverse probability weighting to maximally leverage the observed data and to limit potential bias caused by missing values. The first two of the variants can be viewed as sub-versions of the third and best performing variant, but are important in their own in illustrating the successive improvements in learning accuracy. The empirical investigations show that the proposed approach outperforms the commonly used and state-of-the-art Structural EM algorithm, both in terms of learning accuracy and efficiency, as well as both when data are missing at random and not at random.

\end{abstract}
\section{Introduction}
\label{sec: introduction}
The field of Bayesian Network (BN) structure learning represents a set of approaches that focus on recovering the conditional or causal relationships between variables from data. Structure learning can be divided into two main categories known as constraint-based and score-based methods. Constraint-based methods such as PC~\citep{spirtes2000causation} and IAMB~\citep{tsamardinos2003algorithms} recover a graph by ruling out the structures that violate the conditional independencies discovered from data, and orientating edges by determining colliders. Score-based algorithms such as GES~\citep{chickering2002optimal} and GOBNILP~\citep{cussens2011bayesian} recover a graph by exploring the search space of possible graphs and returning the graph with the highest objective score. While numerous BN structure learning algorithms have been proposed in the literature over the past few decades, most of them do not efficiently learn from data that contain systematic missing values. This hinders the application of structure learning to real-world problems, since missing data represents a common issue in most applied areas including medicine and healthcare~\citep{constantinou2016complex}, clinical epidemiology~\citep{pedersen2017missing}, traffic flow prediction~\citep{tian2018lstm}, anomaly detection~\citep{zemicheal2019anomaly}, and financial analysis~\citep{john2019imputation}. Therefore, there is a greater need for structure learning algorithms that account for potential data bias due to systematic missing values, without having significant impact on the computational efficiency of structure learning.

According to~\cite{rubin1976inference}, missing data problems can be categorised into three classes. These are the Missing Completely At Random (MCAR), the Missing At Random (MAR) and the Missing Not At Random (MNAR). Specifically, MCAR denotes that the missing values are purely random and independent of other observed variables or parameters. This type of missingness is usually caused by technical error that would not bias the analysis. The definition of MAR, on the other hand, is somewhat counterintuitive in its name and assumes the missing values are dependent on observed data. For example, in an investigation between age and frequency of smoking, missing data are MAR if younger respondents are more likely to not disclose their smoking frequency. Lastly, data missingness are said to be MNAR if it is neither MCAR nor MAR. In the above example, the missingness are MNAR if data on respondent's age also contains missing values.

Methods that deal with missing data typically include naïve approaches such as the complete case analysis (a.k.a list-wise deletion) and multiple imputation~\citep{rubin2004multiple}. Complete case analysis involves removing the data cases that contain missing values and hence, restricting learning to complete data cases. Clearly, while this approach is easy to implement, it can be sample inefficient and may yield bias when missingness are not MCAR~\citep{graham2009missing}. Multiple imputation, on the other hand, fills - rather than ignoring - the missing values and takes the uncertainty of imputation into consideration by repeating imputation over different possible values~\citep{azur2011multiple}. However, multiple imputation is built under the assumption of MAR which means it may also produce biased outcomes when data are MNAR.

One of the earliest advanced approaches for dealing with missing data is the Expectation-Maximisation (EM) algorithm, which was also later adopted by the structure learning community. The Structural EM algorithm~\citep{friedman1997learning} is an iterative process which consists of two steps: the Expectation (E) step and the Maximisation (M) step. In E step, Structural EM makes inferences on the missing values and computes the expected sufficient statistics based on the graph learned in previous iteration. The M step follows where the current state of the learned graph is revised based on the sufficient statistics obtained at step E. An advantage of Structural EM is that it can be combined with different structure learning algorithms. A disadvantage, however, is that it is computationally inefficient due to the inference process that takes place at step E. Therefore, in practice, the E step of the Structural EM algorithm is usually implemented with single imputation, i.e., imputing the expectation of the missing values derived from the observed values. \cite{ruggieri2020hard} compared the performance of the original Structural EM to that of the imputed-based Structural EM, and found that the latter achieves better performance in most of the simulation scenarios.

An increasing number of algorithms are recently proposed to improve structure learning from data containing missing values. In the case of score-based learning, two model selection methods have been proposed based on the likelihood function called Node-Average Likelihood (NAL) for discrete~\citep{balov2013consistent} and conditional Gaussian BNs~\citep{bodewes2021learning}. While these methods are consistent with MCAR, they are not consistent with MAR or MNAR cases. In constraint-based learning, \cite{strobl2018fast} treated missing values as a type of selection bias and showed that performing test-wise deletion during conditional independence (CI) tests represents a sound solution for the FCI algorithm~\citep{spirtes2000causation}. In the context of constraint-based learning, test-wise deletion is a process that deletes the data cases with missing values amongst the variables involved in a given CI test. \cite{gain2018structure} later show that replacing the standard CI test in PC with an Inverse Probability Weighting (IPW)~\citep{horvitz1952generalization} based CI test, enables PC to be applied to data sets which contain systematic missing values without loss of consistency. IPW is an approach to alleviate bias in data distributions by reweighting the data cases which we will describe in detail in Section~\ref{sec: handling missing data in hill-climbing algorithm}. However, IPW CI testing assumes sufficient information of missingness, such as information about the parents of missingness and the total ordering of the missing indicators, which is unlikely to be known in practise. \cite{tu2019causal} tried to address this issue by first predicting the parents of missingness using constraint-based learning, for every observed variable that contained missing values, and applying the IPW CI tests using the sufficient information obtained during the constraint-based learning phase.

In this paper, we propose three variants of the greedy search Hill-Climbing algorithm to investigate how they handle missing data values under different assumptions of missingness. These variants can be viewed as fusions between greedy search score-based learning, and the pairwise deletion and IPW methods discussed above that have been previously applied to constraint-based learning. The contribution of this paper is a novel structure learning algorithm suitable for structural learning from data that contain systematic missingness. The empirical results show that, under systematic missingness, the proposed algorithm outperforms the current state-of-the-art Structural EM algorithm, both in terms of learning accuracy and efficiency.

The paper is organised as follows: Section~\ref{sec: preliminaries} provides necessary preliminary information that includes notation and background information, Section~\ref{sec: handling missing data in hill-climbing algorithm} describes the proposed algorithm, Section~\ref{sec: experiments} presents the results, and we provide our concluding remarks in Section~\ref{sec: conclusion}.
\section{Preliminaries}
\label{sec: preliminaries}
In this paper, we consider discrete variables which we denote with uppercase letters (e.g., $U, V$), and the assignment of variable states with lowercase letters (e.g., $u, v$). We denote a set of variables with bold uppercase letters (e.g., $\bm U, \bm V$), and the assignment of a set of variable states with bold lowercase letters (e.g., $\bm u, \bm v$).
\subsection{Bayesian network}
\label{subsec: bayesian network}
A BN $\left<\mathcal{G}, P\right>$ is a probabilistic graphical model that can be represented by a Directed Acyclic Graph (DAG) $\mathcal{G}=\left(\bm V, \bm E\right)$ and a joint distribution $P$ defined over $\bm V$, where $\bm V=\left\{V_1, \ldots, V_n\right\}$ represents a set of random variables and $\bm E$ represents a set of directed edges between pairs of variables. A BN entails the \emph{Markov Condition} which states that for every variable $V_i$ in $\mathcal{G}$, $V_i$ is independent of all its non-descendants conditional on its parents. Given the Markov Condition, the joint distribution $P$ can be factorised as follows:
\begin{equation}
\label{equ: bn}
  P\left(V_1, \ldots, V_n\right) = \prod_{i=1}^n P\left(V_i\mid\bm{Pa}_i\right)\,,
\end{equation}
where $\bm{Pa}_i$ represents the parent-set of $V_i$ in $\mathcal{G}$. Since this study focuses on discrete BNs, we assume that every variable follows an independent multinomial distribution given their parents. We also assume that the set of observed variables $\bm V$ is \emph{causally sufficient}~\citep{spirtes2000causation} and this means that we assume there are no unobserved common causes between any of the variables in $\bm V$. In practice, this means that even though measurement error can be viewed as a hidden variable problem where nodes that contain any form of error must have a hidden parent that causes that error, we assume causal sufficiency such that the graphs reconstructed by SEM are DAGs that contain the observed variables only.

Because an observed distribution can be represented by multiple different DAGs, we work under the assumption that multiple DAGs can be statistically indistinguishable. A collection of DAGs that are statistically indistinguishable, and express the same joint distribution, is also known as a set of Markov equivalent DAGs often referred to as a Completely Partial DAG (CPDAG)~\citep{spirtes2000causation}. A CPDAG can be obtained from a DAG by a) preserving all its v-structures, b) preserving all the directed edges that would create a cycle or a new v-structure if reversed, and c) converting the residual directed edges to undirected edges.
\subsection{Hill Climbing algorithm}
\label{subsec: hill climbing algorithm}
For simplicity, we focus on the Hill-Climbing (HC) structure learning algorithm~\citep{heckerman1995learning} which is a classic score-based learning algorithm that greedily searches the space of neighbouring graphs. It typically starts from an empty graph and explores the search space of graphs via edge additions, deletions and reversals that maximally improve the objective score. HC terminates when no neighbouring graph increases the objective score. HC is an approximate learning algorithm that returns a local maximum solution. However, it is acknowledged to be a computationally efficient algorithm that often outperforms other more complex algorithms~\citep{gamez2011learning, constantinou2021large}. The pseudo-code of the standard HC structure learning algorithm is provided in Algorithm~\ref{alg: hill climbing algorithm}.
\begin{algorithm}[!ht]
\caption{The Hill-Climbing structure learning algorithm}
\label{alg: hill climbing algorithm}
\hspace*{\algorithmicindent} \textbf{Input} data set $D$\\
\hspace*{\algorithmicindent} \textbf{Output} learned DAG $\mathcal{G}$
\begin{algorithmic}[1]
\Procedure{Hill Climbing}{}
    \State $\mathcal{G}\leftarrow$ empty graph
        \Repeat
            \State $\delta\leftarrow 0$
            \Repeat
                \State\parbox[t]{\dimexpr0.9\linewidth-\algorithmicindent}{construct a neighbouring DAG $\mathcal{G}_{nei}$ by \emph{adding}, \emph{reversing} or \emph{deleting} an edge from $\mathcal{G}$}
                \If{$S(\mathcal{G}_{nei}\mid D) - S(\mathcal{G}\mid D) > \delta$}
                    \State $\delta\leftarrow S(\mathcal{G}_{nei}\mid D) - S(\mathcal{G}\mid D)$
                    \State $\mathcal{G}_{update}\leftarrow\mathcal{G}_{nei}$
                \EndIf
            \Until{all possible edge operations have been attempted}
            \If{$\delta > 0$}
                \State $\mathcal{G}\leftarrow\mathcal{G}_{update}$
            \EndIf
    \Until{$\delta=0$}
\EndProcedure
\end{algorithmic}
\end{algorithm}

As with most other structure learning algorithms, HC is usually paired with a decomposable score function to evaluate each graph explored relative to the input data. A score function $S\left(\mathcal{G}, D\right)$ is decomposable if it can be written as the sum over a set of local scores, each of which corresponds to a variable and its parents in $\mathcal{G}$. While all score-based algorithms can use a decomposable score, this property is particular efficient in the case of HC search since it explores one or two graphical modifications at a time; i.e., one in case of edge addition or removal, and two in the case of edge reversal. Therefore, the objective score for each neighbouring graph $\mathcal{G}_{nei}$ can be obtained efficiently by only recomputing the local scores of up to two nodes whose parent-set has changed, and obtaining the local scores of the remaining nodes whose parent-set remains intact from the current best graph $\mathcal{G}$.

Many score functions offer the decomposable property, and most commonly include the \emph{Bayesian Information Criterion} (BIC)~\citep{schwarz1978estimating}, the \emph{Bayesian Dirichlet equivalent} (BDe)~\citep{heckerman1995learning} and the \emph{quotient Normalized Maximum Likelihood} (qNML)~\citep{silander2018quotient}. In this paper, we employ BIC as the score function in all of our experiments. The formal definition of BIC is:
\begin{equation}
\label{equ: bic}
    \begin{split}
        S_{BIC}\left(\mathcal{G}\mid D\right) &= \text{log}L\left(\mathcal{G}\mid D\right) - \frac{\text{log}\left(N\right)}{2}\cdot\lvert\mathcal{G}\rvert \\
         & = \sum_{i=1}^n\left(\text{log}P\left(V_i\mid\bm{Pa}_i, \hat{\Theta}_i\right) - \frac{\text{log}\left(N\right)}{2}\cdot\lvert\hat{\Theta}_i\rvert\right)\,,
    \end{split}
\end{equation}
where $N$ is the sample size, $\bm{Pa}_i$ is the parent set of $V_i$ in $\mathcal{G}$, $\hat{\Theta}_i$ is the maximum likelihood estimates of the parameters over the local distribution of $V_i$, and $\lvert\hat{\Theta}_i\rvert$ is the number of free parameters in $\hat{\Theta}_i$. If $\mathcal{G}$ is defined over a set of discrete multinomial variables $\bm V = \left\{V_1, \ldots V_n\right\}$, then the BIC score has the following form:
\begin{equation}
\label{equ: bic discrete}
    S_{BIC}\left(\mathcal{G}\mid D\right) = \sum_{i=1}^n\left(\sum_{j=1}^{q_i}\sum_{k=1}^{r_i}N_{ijk}\cdot\text{log}\frac{N_{ijk}}{N_{ij}} - \frac{\text{log}\left(N\right)}{2}\cdot (r_i - 1)q_i\right)\,,
\end{equation}
where $N_{ijk}$ is the number of cases in data set $D$ in which the variable $V_i$ takes its $k^{th}$ value and the parents of $V_i$ take the $j^{th}$ configuration. Similarly, $N_{ij}$ is the number of cases in data set $D$ where the parents of $V_i$ take their $j^{th}$ configuration and, therefore, $N_{ij} = \sum_{k=1}^{r_i}N_{ijk}$. Lastly, $r_i$ represents the number of distinct values of $V_i$ and $q_i$ represents the number of configurations of the parents of $V_i$.
\subsection{Missing data assumptions}
\label{subsec: missing data assumptions}
We adopt the graphical descriptions of missing data introduced by~\cite{NIPS2013_0ff8033c} and~\cite{mohan2021graphical}. In this paper, we denote the set of fully observed variables (i.e, variables without missing values) as $\bm V_o$ and the set of partially observed variables (i.e., variables with at least one missing values) as $\bm V_m$. For every partially observed variable $V_i\in \bm V_m$, we define an auxiliary variable $R_i$ called missing indicator to reflect the missingness in $V_i$, where $R_i$ takes the value of 0 when $V_i$ is recorded and the value of 1 when $V_i$ is missing.

Further, we define the \emph{missingness graph} (m-graph~\citep{NIPS2013_0ff8033c}) $\mathcal{G}\left(\mathbb{V}, \bm E\right)$ that captures the relationships between observed variables $\bm V$ and missing indicators $\bm R$, where $\mathbb{V}=\bm V_o\cup\bm V_m\cup\bm R$. Based on m-graph, we define missing data as MCAR if $\bm R\indep\bm V_o\cup\bm V_m$, MAR if $\bm R\indep\bm V_m\mid\bm V_o$, otherwise MNAR. Figure~\ref{fig: m-graph for MCAR, MAR and MNAR} presents the three possible m-graphs assuming three observe variables with structure $V_1\rightarrow V_2\rightarrow V_3$, depicting the MCAR, MAR and MNAR assumptions respectively.
\begin{figure}[!ht]
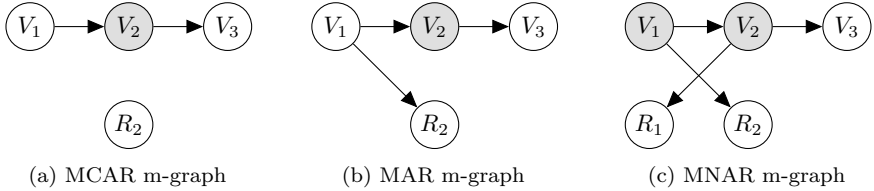

\centering
\begin{subfigure}{.33\textwidth}
  \centering
  \tikz{
    \node[latent, scale = 0.9] (V1) at (-1.3, 0) {$V_1$};
    \node[obs, scale = 0.9] (V2) at (0, 0) {$V_2$};
    \node[latent, scale = 0.9] (V3) at (1.3, 0) {$V_3$};
    \node[latent, scale = 0.9] (R2) at (0, -1.3) {$R_2$};

    \edge {V1} {V2};
    \edge {V2} {V3};
  }
  \caption{MCAR m-graph}
  \label{fig: MCAR}
\end{subfigure}%
\begin{subfigure}{.33\textwidth}
  \centering
  \tikz{
    \node[latent, scale = 0.9] (V1) at (-1.3, 0) {$V_1$};
    \node[obs, scale = 0.9] (V2) at (0, 0) {$V_2$};
    \node[latent, scale = 0.9] (V3) at (1.3, 0) {$V_3$};
    \node[latent, scale = 0.9] (R2) at (0, -1.3) {$R_2$};

    \edge {V1} {V2};
    \edge {V2} {V3};
    \edge {V1} {R2};
  }
  \caption{MAR m-graph}
  \label{fig: MAR}
\end{subfigure}
\begin{subfigure}{.33\textwidth}
  \centering
  \tikz{
    \node[obs, scale = 0.9] (V1) at (-1.3, 0) {$V_1$};
    \node[obs, scale = 0.9] (V2) at (0, 0) {$V_2$};
    \node[latent, scale = 0.9] (V3) at (1.3, 0) {$V_3$};
    \node[latent, scale = 0.9] (R1) at (-1.3, -1.3) {$R_1$};
    \node[latent, scale = 0.9] (R2) at (0, -1.3) {$R_2$};

    \edge {V1} {V2};
    \edge {V2} {V3};
    \edge {V1} {R2};
    \edge {V2} {R1};
  }
  \caption{MNAR m-graph}
  \label{fig: MNAR}
\end{subfigure}
\caption{The three possible m-graphs assuming three observed variables with structure $V_1\rightarrow V_2\rightarrow V_3$. Shaded nodes represent partially observed variables.}
\label{fig: m-graph for MCAR, MAR and MNAR}
\end{figure}

To ensure the population distributions are recoverable from the observed data, some assumptions need to be employed for the missing indicators. These are:
\begin{assumption}
\label{ass: r is effect node}
Variables in $\bm R$ neither can be the parent of an observed variables in $\bm V$ nor other variables in $\bm R$.
\end{assumption}
\begin{assumption}
\label{ass: no cause between variable and its own indicator}
No partially observed variable can be the parent of its own missing indicator.
\end{assumption}
Assumption~\ref{ass: r is effect node} states that a missing indicator in $\bm R$ can only be an effect (leaf) node in an m-graph, whereas Assumption~\ref{ass: no cause between variable and its own indicator} states that the missing value is independent of the variable value. When both Assumption~\ref{ass: r is effect node} and~\ref{ass: no cause between variable and its own indicator} hold, the joint distribution of the observed variables is recoverable from the observed data~\cite[Theorem~2]{NIPS2013_0ff8033c}.
\section{Handling systematic missing data with Hill-Climbing}
\label{sec: handling missing data in hill-climbing algorithm}
This section describes the three HC variants that we explore in extending the learning process towards dealing with systematic missing data. Specifically, subsection~\ref{subsec: hill-climbing with pairwise deletion} describes HC with pairwise deletion which we call HC-pairwise, subsection~\ref{subsec: hill-climbing with inverse probability weighting} describes HC with both pairwise deletion and Inverse Probability Weighting which we call HC-IPW, and subsection~\ref{subsec: hill-climbing with adaptive inverse probability weighting} describes an improved version of HC-IPW, the HC-aIPW, that prunes less data samples compared to HC-IPW. The first two HC-variants can be viewed as sub-versions of HC-aIPW, but are important in their own in illustrating the successive improvements in learning accuracy.
\subsection{Hill-Climbing with pairwise deletion}
\label{subsec: hill-climbing with pairwise deletion}
Recall that, at each iteration, HC moves to the neighbouring graph that maximally improves the objective score, and that performing HC search with a decomposable scoring function means that there is no need to recompute the local score of variables whose parent-set remains unchanged across graphs. Therefore, an efficient (but not necessarily effective) way of applying HC to missing data is to ignore data cases that contain missing values in variables that form part of the set of variables considered when exploring local score changes to a DAG. We refer to this process as \emph{pairwise deletion}, where ``pair'' refers to the current pair of candidate DAGs (the current best DAG and neighbouring DAG), and this deletion process may involve more than two variables. When comparing the current best DAG against a neighbouring DAG, the \emph{necessary variables} would be the nodes with unequal parent-sets between the two graphs, plus the parents of those nodes in the two graphs. Formally, when exploring a neighbouring DAG $\mathcal{G}_{nei}$ from the current best DAG $\mathcal{G}$, the set of necessary variables $\bm W$ between $\mathcal{G}$ and $\mathcal{G}_{nei}$ can be described as:
\begin{equation}
    \label{equ: necessary variables}
    \bm W=\cup_{V_i\in\bm V_d}\left\{V_i, \bm{Pa}_i, \bm{Pa}_i^{nei}\right\}\,,
\end{equation}
where $\bm V_d$ is the set of variables that have different parent-sets between $\mathcal{G}$ and $\mathcal{G}_{nei}$, and $\bm{Pa}_i$ and $\bm{Pa}_i^{nei}$ are the parent-sets of $V_i$ in $\mathcal{G}$ and $\mathcal{G}_{nei}$ respectively. For simplicity, we refer to the data set obtained after applying pairwise deletion as the pairwise deleted data set.
\begin{example}
\label{exa: necessary variables}
Assume that, during HC, the current state of DAG $\mathcal{G}$ is a graph containing three variables $\left\{V_1, V_2, V_3\right\}$ and the edge $V_1\rightarrow V_2$, as illustrated in Table~\ref{tab: examples of necessary variables}. Given DAG $\mathcal{G}$, there are six possible edge operations each of which produces a neighbouring graph $\mathcal{G}_{nei}$. Operation add $V_1\rightarrow V_3$, for example, can be evaluated by assessing the change in the local score of $V_3$, i.e., $S\left(V_3\mid V_1\right) - S\left(V_3\right)$, since $V_3$ is the only variable with different parents between $\mathcal{G}$ and $\mathcal{G}_{nei}$. When the data set contains missing values, we can apply pairwise deletion to data given $\left\{V_1, V_3\right\}$ in order to obtain a complete data set that will enable us to assess the neighbouring graph resulting from this edge operation. However, there is a risk that this action may lead to biased estimates when missingness is not MCAR.
\end{example}
\begin{table}[!ht]
\centering
\caption{Examples of necessary variables for each edge operation in HC, which we define as the variables with different parent-sets between the current best and neighbouring graphs, plus the parents that make up those parent-sets.}
\begin{tabular}{cccc}
\toprule
current DAG state $\mathcal{G}$ & edge operation & neighbouring DAG $\mathcal{G}_{nei}$ & necessary variables\\
\midrule
\multirow{6}{*}[-1.7cm]{\begin{tikzpicture}[baseline=0]
    \node[latent, scale = 0.8] (V1) at (-1.1, 0) {$V_1$};
    \node[latent, scale = 0.8] (V2) at (0, 0) {$V_2$};
    \node[latent, scale = 0.8] (V3) at (1.1, 0) {$V_3$};
    \edge {V1} {V2};
  \end{tikzpicture}} & \multirow{1}{*}[-1.5mm]{add $V_1\rightarrow V_3$} & \begin{tikzpicture}[baseline=0]
    \node[latent, scale = 0.8] (V1) at (-1.1, 0) {$V_1$};
    \node[latent, scale = 0.8] (V2) at (0, 0) {$V_2$};
    \node[latent, scale = 0.8] (V3) at (1.1, 0) {\small $V_3$};
    \edge {V1} {V2};
    \path [->] (V1) edge [bend right] (V3);
  \end{tikzpicture} & \multirow{1}{*}[-1.5mm]{$\left\{V_1, V_3\right\}$}\\\cmidrule{2-4}
& \multirow{1}{*}[-1mm]{add $V_2\rightarrow V_3$} & \begin{tikzpicture}[baseline=0]
    \node[latent, scale = 0.8] (V1) at (-1.1, 0) {$V_1$};
    \node[latent, scale = 0.8] (V2) at (0, 0) {$V_2$};
    \node[latent, scale = 0.8] (V3) at (1.1, 0) {$V_3$};
    \edge {V1} {V2};
    \edge {V2} {V3};
    \addvmargin{1mm}
  \end{tikzpicture} & \multirow{1}{*}[-1mm]{$\left\{V_2, V_3\right\}$}\\\cmidrule{2-4}
& \multirow{1}{*}[-1.5mm]{add $V_3\rightarrow V_1$} &\begin{tikzpicture}[baseline=0]
    \node[latent, scale = 0.8] (V1) at (-1.1, 0) {$V_1$};
    \node[latent, scale = 0.8] (V2) at (0, 0) {$V_2$};
    \node[latent, scale = 0.8] (V3) at (1.1, 0) {$V_3$};
    \edge {V1} {V2};
    \path [->] (V3) edge [bend left] (V1);
  \end{tikzpicture} & \multirow{1}{*}[-1.5mm]{$\left\{V_1, V_3\right\}$}\\\cmidrule{2-4}
& \multirow{1}{*}[-1mm]{add $V_3\rightarrow V_2$} &\begin{tikzpicture}[baseline=0]
    \node[latent, scale = 0.8] (V1) at (-1.1, 0) {$V_1$};
    \node[latent, scale = 0.8] (V2) at (0, 0) {$V_2$};
    \node[latent, scale = 0.8] (V3) at (1.1, 0) {$V_3$};
    \edge {V1} {V2};
    \edge {V3} {V2};
    \addvmargin{1mm}
  \end{tikzpicture} & \multirow{1}{*}[-1mm]{$\left\{V_1, V_2, V_3\right\}$}\\\cmidrule{2-4}
& \multirow{1}{*}[-1mm]{reverse $V_1\rightarrow V_2$} &\begin{tikzpicture}[baseline=0]
    \node[latent, scale = 0.8] (V1) at (-1.1, 0) {$V_1$};
    \node[latent, scale = 0.8] (V2) at (0, 0) {$V_2$};
    \node[latent, scale = 0.8] (V3) at (1.1, 0) {$V_3$};
    \edge {V2} {V1};
    \addvmargin{1mm}
  \end{tikzpicture} & \multirow{1}{*}[-1mm]{$\left\{V_1, V_2\right\}$}\\\cmidrule{2-4}
& \multirow{1}{*}[-1mm]{delete $V_1\rightarrow V_2$} &\begin{tikzpicture}[baseline=0]
    \node[latent, scale = 0.8] (V1) at (-1.1, 0) {$V_1$};
    \node[latent, scale = 0.8] (V2) at (0, 0) {$V_2$};
    \node[latent, scale = 0.8] (V3) at (1.1, 0) {$V_3$};
    \addvmargin{1mm}
  \end{tikzpicture} & \multirow{1}{*}[-1mm]{$\left\{V_1, V_2\right\}$}\\
\bottomrule
\end{tabular}
\label{tab: examples of necessary variables}
\end{table}%

Because pairwise deletion leads to edge operations that are assessed based on different subsets of the data, it is possible to get stuck in an infinite loop where previous neighbouring graphs are constantly revisited and re-selected as a higher scoring graph. This can happen when, for example, DAG $\mathcal{G}_2$ returns a higher score than $\mathcal{G}_1$ based on pairwise deleted data set $D_1$, $\mathcal{G}_3$ returns a higher score than $\mathcal{G}_2$ based on pairwise deleted data set $D_2$, and $\mathcal{G}_1$ returns a higher score than $\mathcal{G}_3$ based on pairwise deleted data set $D_3$. In this example, HC with pairwise deletion would identify the graphical scores as $\mathcal{G}_1<\mathcal{G}_2<\mathcal{G}_3<\mathcal{G}_1$ and never converge to a maximal solution. We address this issue by restricting HC search to neighbours not previously identified as the optimal graph. We call this variant of HC as HC-pairwise, and present its pseudo-code in Algorithm~\ref{alg: hc-pairwise algorithm}.
\begin{algorithm}[!ht]
    \caption{HC-pairwise algorithm}\label{alg: hc-pairwise algorithm}
    \hspace*{\algorithmicindent} \textbf{Input} data set $D$\\
    \hspace*{\algorithmicindent} \textbf{Output} learned DAG $\mathcal{G}$
    \begin{algorithmic}[1]
        \Procedure{HC-pairwise}{}
        \State $\mathcal{G}\leftarrow$ empty graph
        \State $\mathcal{G}_{record}\leftarrow\left\{\mathcal{G}\right\}$
        \Repeat
            \State $\delta\leftarrow 0$
            \Repeat
                \State\parbox[t]{\dimexpr0.9\linewidth-\algorithmicindent}{construct a neighbouring DAG $\mathcal{G}_{nei}$ by \emph{adding}, \emph{reversing} or \emph{deleting} an edge from $\mathcal{G}$}
                \If{$\mathcal{G}_{nei}\not\in\mathcal{G}_{record}$}
                    \State construct $D_{pw}$ by pairwise deleting $D$ given the necessary variables $\bm W$
                    \If{$S(\mathcal{G}_{nei}\mid D_{pw}) - S(\mathcal{G}\mid D_{pw}) > \delta$}
                        \State $\delta\leftarrow S(\mathcal{G}_{nei}\mid D_{pw}) - S(\mathcal{G}\mid D_{pw})$
                        \State $\mathcal{G}_{update}\leftarrow\mathcal{G}_{nei}$
                    \EndIf
                \EndIf
            \Until{all possible edge operations have been attempted}
            \If{$\delta > 0$}
                \State $\mathcal{G}\leftarrow\mathcal{G}_{update}$
                \State $\mathcal{G}_{record}=\mathcal{G}_{record}\cup\left\{\mathcal{G}\right\}$
            \EndIf
        \Until{$\delta=0$}
        \EndProcedure
    \end{algorithmic}
\end{algorithm}

When data are MCAR, on the basis of $\bm R\indep\bm V$ the distribution entailed by any pairwise deleted data set is an unbiased estimate of the underlying true distribution:
\begin{equation}
\label{equ: local MCAR}
    P\left(V_i\mid\bm{Pa}_i, \bm R_s=\bm 0\right) = P\left(V_i\mid\bm{Pa}_i\right)\,,
\end{equation}
where $\bm R_s$ can be any subset of $\bm R$.

From this, we derive Proposition~\ref{pro: mcar}, which states that, when the missingness is MCAR, the DAG learned by HC-pairwise is a local maximum graph, at least when BIC is used as the objective function. We define the local maximum graph as the graph with an objective score not lower than the scores of all its valid neighbouring graphs, when these scores are derived from the fully observed data set; i.e., it is independent of missingness generated.
\begin{restatable}{prop}{mcar}
\label{pro: mcar}
Assume data $D$ is MCAR and sample size $N\rightarrow\infty$, for any DAG $\mathcal{G}$ and one of its neighbouring DAG $\mathcal{G}_{nei}$
\begin{equation*}
    S_{BIC}\left(\mathcal{G}_{nei}\mid D_{pw}\right) > S_{BIC}\left(\mathcal{G}\mid D_{pw}\right) \text{, iff } S_{BIC}\left(\mathcal{G}_{nei}\mid D_{f}\right) > S_{BIC}\left(\mathcal{G}\mid D_{f}\right)\,,
\end{equation*}
where $D_{pw}$ is the pairwise deleted data set which is derived from $D$ by removing the data cases with missing values amongst the necessary variables $\bm W$, and $D_f$ is the corresponding fully observed data set.
\end{restatable}
\subsection{Hill-Climbing with Inverse Probability Weighting}
\label{subsec: hill-climbing with inverse probability weighting}
Although HC-pairwise will progressively learn a better DAG after each iteration when missingness is MCAR, this property does not necessarily hold when missingness is MAR or MNAR, since systematic bias in the data might produce
\begin{equation}
    \label{equ: mnar}
    P\left(V_i\mid\bm{Pa}_i, \bm R=\bm 0\right)\neq P\left(V_i\mid\bm{Pa}_i\right)\,.
\end{equation}
To diminish data biases caused by potential dependencies between missing and observed data, we further explore applying the IPW method on the pairwise deleted data set.

According to~\citet[Theorem~2]{NIPS2013_0ff8033c} and~\cite{tu2019causal}, when Assumptions~\ref{ass: r is effect node} and~\ref{ass: no cause between variable and its own indicator} hold, the joint distribution of variables $\bm V$ can be fully recovered from the observed part of the data set (i,e., the data after applying pairwise deletion) by
\begin{multline}
    \label{equ: recover function}
    P\left(\bm V\right) = P\left(\bm V\mid\bm R=\bm 0\right)\cdot\\\underbrace{\frac{P\left(\bm R=\bm 0\right)}{\prod_{R_i\in\bm R}P\left(R_i=0\mid\bm R_{\bm{Pa}_{R_i}}=\bm 0\right)}}_c\prod_{R_i\in\bm R}\underbrace{\frac{P\left(\bm{Pa}_{R_i}\mid\bm R_{\bm{Pa}_{R_i}}=\bm 0\right)}{P\left(\bm{Pa}_{R_i}\mid R_i = 0,\bm R_{\bm{Pa}_{R_i}}=\bm 0\right)}}_{\beta_{R_i}}\,,
\end{multline}
where $\bm{Pa}_{R_i}$ is the set of parents of missing indicator $R_i$, and $\bm R_{\bm{Pa}_{R_i}}$ is the set of missing indicator of the partially observed variables in $\bm{Pa}_{R_i}$. We further discuss and provide the derivation of Equation~\ref{equ: recover function} in Appendix~\ref{app: derivation of equation 7}.

Since the term $c$ in Equation~\ref{equ: recover function} represents a constant value, we can apply pairwise deletion to the missing data cases of variables $\bm V$ and weight the pairwise deleted data set by $\prod_{Ri\in R}\beta_{R_i}$. This will produce a weighted data set that approximates the unbiased distribution $P\left(\bm V\right)$. We call this HC variant HC-IPW, and can be viewed as an extension of HC-pairwise that incorporates both the pairwise deletion and IPW methods. Unlike HC-pairwise, the HC-IPW algorithm can be used under the assumption the input data are MAR or MNAR, in addition to MCAR, to diminish data bias caused by systematic missing values.

It should be noted that when $\bm{Pa}_{R_i}$ contains partially observed variables, Equation~\ref{equ: recover function} implies that $\bm{Pa}_{R_i}\subseteq\bm V$; otherwise, the columns of $\bm{Pa}_{R_i}$ in the pairwise deleted data set may contain missing values that will render the calculation of $\beta_{R_i}$ invalid. The following example shows that it might be impossible to recover the underlying true distribution if any $\bm{Pa}_{R_i}\not\subseteq\bm V$.
\begin{example}
\label{exa: issue of using w in ipw}
Consider that Figure~\ref{fig: MNAR} is the true m-graph, the current best DAG $\mathcal{G}$ in HC search is the one shown in Figure~\ref{fig: current best dag}, and Figure~\ref{fig: neighbouring dag} presents one of its neighbouring DAGs, $\mathcal{G}_{nei}$. Since the difference in score between $\mathcal{G}_{nei}$ and $\mathcal{G}$ is $S\left(V_3\mid V_1\right) - S\left(V_3\right)$, we need to ensure that missingness does not bias the estimate of distribution $P\left(V_1, V_3\right)$ when computing distributional score difference. If we apply pairwise deletion directly on the necessary variables $\left\{V_1, V_3\right\}$ and use Equation~\ref{equ: recover function} to recover $P\left(V_1, V_3\right)$. This will result in the following equation:
\begin{equation*}
    P\left(V_1, V_3\right) = P\left(V_1, V_3\mid R_1 = 0\right)\frac{P\left(R_1 = 0\right)}{P\left(R_1 = 0\right)}\cdot\frac{P\left(V_2\mid R_2 = 0\right)}{P\left(V_2\mid R_1 = 0, R_2 = 0\right)}\,.
\end{equation*}
However, the problem in the above equation is that we cannot compute the weight term $\frac{P\left(V_2\mid R_2 = 0\right)}{P\left(V_2\mid R_1 = 0, R_2 = 0\right)}$ for data cases that contain missing values in $V_2$.
\begin{figure}[!ht]
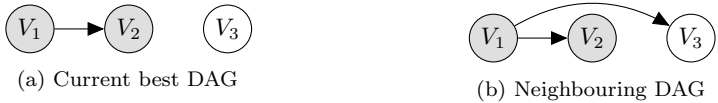

\centering
\begin{subfigure}{.5\textwidth}
  \centering
  \tikz{
    \node[obs, scale = 0.9] (V1) at (-1.3, 0) {$V_1$};
    \node[obs, scale = 0.9] (V2) at (0, 0) {$V_2$};
    \node[latent, scale = 0.9] (V3) at (1.3, 0) {$V_3$};

    \edge {V1} {V2};
  }
  \caption{Current best DAG}
  \label{fig: current best dag}
\end{subfigure}%
\begin{subfigure}{.5\textwidth}
  \centering
  \tikz{
    \node[obs, scale = 0.9] (V1) at (-1.3, 0) {$V_1$};
    \node[obs, scale = 0.9] (V2) at (0, 0) {$V_2$};
    \node[latent, scale = 0.9] (V3) at (1.3, 0) {$V_3$};

    \edge {V1} {V2};
    \path [->] (V1) edge [bend left] (V3);
  }
  \caption{Neighbouring DAG}
  \label{fig: neighbouring dag}
\end{subfigure}
\caption{A hill-climbing illustration of the DAG considered in Example~\ref{exa: issue of using w in ipw}, discussed in the main text. Shaded nodes represent partially observed variables.}
\label{fig: example of recovery function}
\end{figure}
\end{example}
To avoid this, when assessing the edge operations from $\mathcal{G}$ to $\mathcal{G}_{nei}$ in HC-IPW, the pairwise deletion for Equation~\ref{equ: recover function} should be performed on \emph{sufficient variables} $\bm U$, which is a variable set that contains the necessary variables $\bm W$ plus the parents of missing indicators of all variables in $\bm U$:
\begin{equation}
    \label{equ: sufficient variables}
    \bm U=\cup_{V_i\in\bm V_d}\left\{V_i, \bm{Pa}_i, \bm{Pa}_i^{nei}\right\}\cup\bm{Pa}_{\bm R_{\bm U}}\,,
\end{equation}
where $\bm V_d$ is the set of variables that have different parent-sets between $\mathcal{G}$ and $\mathcal{G}_{nei}$, and $\bm{Pa}_i$ and $\bm{Pa}_i^{nei}$ are the parent-sets of $V_i$ in $\mathcal{G}$ and $\mathcal{G}_{nei}$ respectively. It is worth noting that Equation\mbox{~\ref{equ: sufficient variables}} represents a recursive process that iterates over the parents of missing indicators for all involved variables, i.e., not only $\bm W$ but also $\bm U\backslash\bm W$ should be included in $\bm U$ in order to resolve the issue illustrated in Example\mbox{~\ref{exa: issue of using w in ipw}}.

Another potential issue with Equation~\ref{equ: recover function} is that the parents $\bm{Pa}_{R_i}$ of each missing indicator $R_i$ are generally unknown. \cite{tu2019causal} used constraint-based learning to discover the parents of each missing indicator, and this approach has been proven to be sound when both Assumptions~\ref{ass: r is effect node} and~\ref{ass: no cause between variable and its own indicator} hold. We have, therefore, adopted the constraint-based approach proposed by~\cite{tu2019causal} to discover the parents of the missing indicators in applying HC-IPW. The intention here is that this approach can be used to exclude variable $V_j$ as the parent of $R_i$, if $R_i$ is found to be independent of $V_j$ given any variable set $\bm S$, given the pairwise deleted data set for $\left\{V_j\right\}\cup\bm S$. Algorithm~\ref{alg: direct cause detection} provides the pseudo-code.
\begin{algorithm}[!ht]
    \caption{Discovering the parents of the missing indicators using constraint-based learning}\label{alg: direct cause detection}
    \hspace*{\algorithmicindent} \textbf{Input} data set $D$\\
    \hspace*{\algorithmicindent} \textbf{Output} the parents of missing indicators $\bm{Pa}_{\bm R}$
    \begin{algorithmic}[1]
        \Procedure{Detecting parents of missing indicators}{}
        \ForEach{$V_i\in\bm V_m$}
            \State $\bm{Pa}_{R_i}\leftarrow\bm V\backslash V_i$
            \ForEach{$V_j\in\bm V\backslash V_i$}
                \State remove $V_j$ from $\bm{Pa}_{R_i}$ if $R_i\indep V_j\mid\bm S, R_j = 0, \bm R_{\bm S} = \bm 0$, for any $\bm S\subseteq\bm{Pa}_{R_i}$
            \EndFor
        \EndFor
        \Return $\bm{Pa}_{\bm R}$
        \EndProcedure
    \end{algorithmic}
\end{algorithm}
\begin{algorithm}[!ht]
    \caption{HC-IPW algorithm}\label{alg: hc-ipw algorithm}
    \hspace*{\algorithmicindent} \textbf{Input} data set $D$\\
    \hspace*{\algorithmicindent} \textbf{Output} learned DAG $\mathcal{G}$
    \begin{algorithmic}[1]
        \Procedure{HC-IPW}{}
        \State $\mathcal{G}\leftarrow$ empty graph
        \State $\mathcal{G}_{record}\leftarrow\left\{\mathcal{G}\right\}$
        \State \textcolor{blue}{retrieve the parents of missing indicators via Algorithm~\ref{alg: direct cause detection}}
        \Repeat
            \State $\delta\leftarrow 0$
            \Repeat
                \State\parbox[t]{\dimexpr0.9\linewidth-\algorithmicindent}{construct a neighbouring DAG $\mathcal{G}_{nei}$ by \emph{adding}, \emph{reversing} or \emph{deleting} an edge from $\mathcal{G}$}
                \If{$\mathcal{G}_{nei}\not\in\mathcal{G}_{record}$}
                    \State construct $D_{pw}$ by pairwise deleting $D$ given the \textcolor{blue}{sufficient variables $\bm U$}
                    \State \textcolor{blue}{compute weight $\beta$ by Equation~\ref{equ: recover function} for $D_{pw}$}
                    \If{$S(\mathcal{G}_{nei}\mid D_{pw}, \textcolor{blue}{\beta}) - S(\mathcal{G}\mid D_{pw}, \textcolor{blue}{\beta}) > \delta$}
                    \State $\delta\leftarrow S(\mathcal{G}_{nei}\mid D_{pw}, \textcolor{blue}{\beta}) - S(\mathcal{G}\mid D_{pw}, \textcolor{blue}{\beta})$
                    \State $\mathcal{G}_{update}\leftarrow\mathcal{G}_{nei}$
                    \EndIf
                \EndIf
            \Until{all possible edge operations have been attempted}
            \If{$\delta > 0$}
                \State $\mathcal{G}\leftarrow\mathcal{G}_{update}$
                \State $\mathcal{G}_{record}=\mathcal{G}_{record}\cup\left\{\mathcal{G}\right\}$
            \EndIf
        \Until{$\delta=0$}
        \EndProcedure
    \end{algorithmic}
\end{algorithm}

Algorithm~\ref{alg: hc-ipw algorithm} describes the HC-IPW algorithm, where lines coloured in blue represent the difference in pseudo-code between HC-IPW and HC-pairwise. Note that when computing the objective score for HC-IPW, the weighted statistics $\widetilde{N}_{ijk}, \widetilde{N}_{ij}$ are used instead of the standard $N_{ijk}, N_{ij}$ used in HC, and which are defined as follows:
\begin{equation}
\label{equ: nijk}
    \widetilde{N}_{ijk} = \sum_{s=1}^{\mid D_{pw}\mid} 1_{ijk}\left(d^s\right)\cdot\beta^s\,,
\end{equation}
\begin{equation}
\label{equ: nij}
    \widetilde{N}_{ij} = \sum_{k=1}^{r_i}\widetilde{N}_{ijk}\,,
\end{equation}
where $1_{ijk}$ is the indicator function of the event $\left(V_i=k, \bm{Pa}_i=j\right)$ which returns 1 when the combination of $V_i=k, \bm{Pa}_i=j$ appears in the input data case, and returns 0 otherwise, $d^s$ is the $s^{th}$ record in pairwise deleted data set $D_{pw}$, and $\beta^s$ is the weight corresponding to $d^s$. Therefore, we define the BIC score for pairwise deleted data set $D_{pw}$ given $\beta$ as follows:
\begin{equation*}
    S_{BIC}\left(\mathcal{G}\mid D_{pw}, \beta\right) = \sum_{i=1}^n\left(\sum_{j=1}^{q_i}\sum_{k=1}^{r_i}\widetilde{N}_{ijk}\cdot\text{log}\frac{\widetilde{N}_{ijk}}{\widetilde{N}_{ij}} - \frac{\text{log}\left(N_{pw}\right)}{2}\cdot (r_i - 1)q_i\right)\,,
\end{equation*}
where $N_{pw}$ represents the sample size of $D_{pw}$, $\widetilde{N}_{ijk}$ and $\widetilde{N}_{ij}$ represent the weighted statistics as defined in Equation\mbox{~\ref{equ: nijk}} and\mbox{~\ref{equ: nij}}, and $\beta$ is used for computing the weighted $\widetilde{N}_{ijk}$ and $\widetilde{N}_{ij}$.

The following proposition shows that HC-IPW converges to a local optima when BIC is used as the score function, when both Assumptions~\ref{ass: r is effect node} and~\ref{ass: no cause between variable and its own indicator} hold, and when sample size $N\rightarrow\infty$.
\begin{restatable}{prop}{mnar}
\label{pro: mnar}
Given Assumptions~\ref{ass: r is effect node} and~\ref{ass: no cause between variable and its own indicator}, assume data $D$ is partially observed and sample size $N\rightarrow\infty$, for any DAG $\mathcal{G}$ and one of its neighbouring DAG $\mathcal{G}_{nei}$
\begin{equation*}
    S_{BIC}\left(\mathcal{G}_{nei}\mid D_{pw}, \beta\right) > S_{BIC}\left(\mathcal{G}\mid D_{pw}, \beta\right)\text{, iff }S_{BIC}\left(\mathcal{G}_{nei}\mid D_f\right) > S_{BIC}\left(\mathcal{G}\mid D_{f}\right)\,,
\end{equation*}
where $D_{pw}$ is the pairwise deleted data set which is derived from $D$ by removing data cases with missing values among sufficient variables $\bm U$, $\beta = \prod_{R_i\in\bm R_{\bm U}}\beta_{R_i}$, and $D_f$ is the corresponding fully observed data set.
\end{restatable}
\subsection{Hill-Climbing with adaptive Inverse Probability Weighting}
\label{subsec: hill-climbing with adaptive inverse probability weighting}
Although HC-IPW diminishes potential data bias caused by systematic missing values, the learning approach achieves this by removing a greater number of data cases compared to those removed by HC-pairwise when $\bm{Pa}_{\bm R_{\bm W}}$ contains partially observed variables, which is likely to happen when the missingness are MNAR. This can be a problem when data cases are limited. We illustrate this phenomenon with an example.
\begin{example}
Suppose graph (a) in Figure~\ref{fig: example of a searching step in HC variants} represents the ground truth m-graph in which the variables in shaded backcolour $V_1, V_4$ and $V_6$ are partially observed whose missingness are caused by $V_4, V_5$ and $V_1$ respectively, as illustrated with the missing indicators $R_1, R_4$ and $R_6$ corresponding to the missingness of $V_1, V_4$ and $V_6$. Let us assume graph (b) represents the current state of the optimal DAG in the HC-pairwise/HC-IPW search process, and that graphs (c) and (d) represent two of the possible neighbouring graphs. When HC-pairwise compares $\mathcal{G}$ with $\mathcal{G}_{n1}$, it applies pairwise deletion on cases in which the necessary variables $\bm W = \left\{V_5, V_2, V_6\right\}$ contain missing values. Since only $V_6$ is partially observed out of the three necessary variables, HC-pairwise removes data cases when the value of $V_6$ is missing. In contrast, when HC-IPW is applied to this case, and assuming it correctly learns the parents of missingness via Algorithm~\ref{alg: direct cause detection}, it computes the weights of the pairwise deleted data set through pairwise deletion based on the sufficient variables $\bm U = \left\{V_5, V_2, V_6\right\}\cup\left\{V_1, V_4, V_5\right\}$. Thus, HC-IPW removes data cases whenever any of the variables in $U$ contain a missing value (in this example, $V_1, V_4$ and $V_6$ do). Therefore, HC-IPW performs learning on a smaller set of data cases compared to those in the case of HC-pairwise.

When $\bm{Pa}_{\bm R_{\bm W}}$ (refer to Equation~\ref{equ: necessary variables} and Algorithm~\ref{alg: hc-ipw algorithm}) does not contain any partially observed variables, the HC-IPW algorithm will perform learning on the same number of data cases as in HC-pairwise. This can happen in cases such as when comparing neighbouring DAG $\mathcal{G}_{n2}$ against $\mathcal{G}$ in Figure~\ref{fig: example of a searching step in HC variants}, where the set of necessary variables $\bm W$ in HC-pairwise contains $\left\{V_4, V_2, V_5\right\}$ and the set of sufficient variables $\bm U$ in HC-IPW is $\left\{V_4, V_2, V_5\right\}\cup\left\{V_5\right\}$. In this case, because $V_5$ is fully observed, applying pairwise deletion given $\bm W$ and $\bm U$ would result in the same pairwise deleted data set.
\end{example}
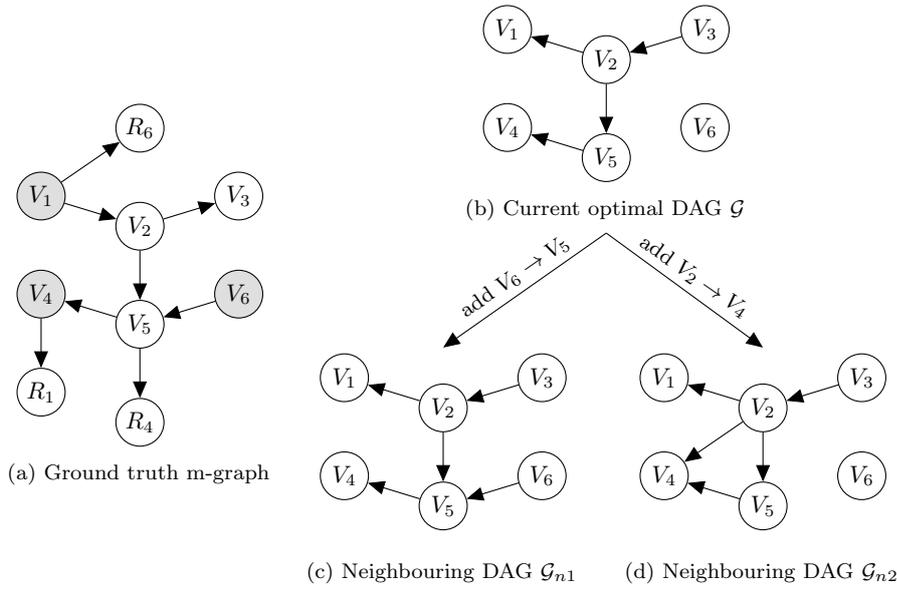
\begin{figure}[!ht]
    \centering
    \begin{minipage}{.3\linewidth}
        \begin{subfigure}{\linewidth}
          \centering
          \tikz{
            \node [obs, scale = 0.9] (V1) at (-1.3, 0) {$V_1$};
            \node [latent, scale = 0.9] (V2) at (0, -0.4) {$V_2$};
            \node [latent, scale = 0.9] (V3) at (1.3, 0) {$V_3$};
            \node [obs, scale = 0.9] (V4) at (-1.3, -1.3) {$V_4$};
            \node [latent, scale = 0.9] (V5) at (0, -1.7) {$V_5$};
            \node [obs, scale = 0.9] (V6) at (1.3, -1.3) {$V_6$};
            \node [latent, scale = 0.9] (R1) at (-1.3, -2.6) {$R_1$};
            \node [latent, scale = 0.9] (R4) at (0, -3) {$R_4$};
            \node [latent, scale = 0.9] (R6) at (0, 0.9) {$R_6$};
            \edge {V1} {V2};
            \edge {V2} {V3};
            \edge {V5} {V4};
            \edge {V6} {V5};
            \edge {V2} {V5};
            \edge {V1} {R6};
            \edge {V4} {R1};
            \edge {V5} {R4};
          }
          \caption{Ground truth m-graph}
          \label{fig: ground truth m-graph}
        \end{subfigure}
    \end{minipage}
    \begin{minipage}{.69\linewidth}
        \begin{subfigure}[t]{\linewidth}
            \centering
            \tikz[remember picture]{
                \node [latent, scale = 0.9] (V1) at (-1.3, 0) {$V_1$};
                \node [latent, scale = 0.9] (V2) at (0, -0.4) {$V_2$};
                \node [latent, scale = 0.9] (V3) at (1.3, 0) {$V_3$};
                \node [latent, scale = 0.9] (V4) at (-1.3, -1.3) {$V_4$};
                \node [latent, scale = 0.9] (V5) at (0, -1.7) {$V_5$};
                \node [latent, scale = 0.9] (V6) at (1.3, -1.3) {$V_6$};
                \edge {V2} {V1};
                \edge {V3} {V2};
                \edge {V5} {V4};
                \edge {V2} {V5};
                \node[coordinate] (A) at (0, -2.7) {};
            }
            \caption{Current optimal DAG $\mathcal{G}$}
            \label{fig: current optimal dag}
        \end{subfigure}\vspace*{1.5cm} \\
        \begin{minipage}{.49\linewidth}
            \begin{subfigure}[b]{\linewidth}
                \centering
                \tikz[remember picture]{
                    \node [latent, scale = 0.9] (V1) at (-1.3, 0) {$V_1$};
                    \node [latent, scale = 0.9] (V2) at (0, -0.4) {$V_2$};
                    \node [latent, scale = 0.9] (V3) at (1.3, 0) {$V_3$};
                    \node [latent, scale = 0.9] (V4) at (-1.3, -1.3) {$V_4$};
                    \node [latent, scale = 0.9] (V5) at (0, -1.7) {$V_5$};
                    \node [latent, scale = 0.9] (V6) at (1.3, -1.3) {$V_6$};
                    \edge {V2} {V1};
                    \edge {V3} {V2};
                    \edge {V5} {V4};
                    \edge {V2} {V5};
                    \edge {V6} {V5};
                    \node[coordinate] (B) at (0, 0.4) {};
                    \addvmargin{2mm};
                }
                \caption{Neighbouring DAG $\mathcal{G}_{n1}$}
                \label{fig: neighbouring dag 1}
            \end{subfigure} 
        \end{minipage}
        \begin{minipage}{.49\linewidth}
            \begin{subfigure}[b]{\linewidth}
                \centering
                \tikz[remember picture]{
                    \node [latent, scale = 0.9] (V1) at (-1.3, 0) {$V_1$};
                    \node [latent, scale = 0.9] (V2) at (0, -0.4) {$V_2$};
                    \node [latent, scale = 0.9] (V3) at (1.3, 0) {$V_3$};
                    \node [latent, scale = 0.9] (V4) at (-1.3, -1.3) {$V_4$};
                    \node [latent, scale = 0.9] (V5) at (0, -1.7) {$V_5$};
                    \node [latent, scale = 0.9] (V6) at (1.3, -1.3) {$V_6$};
                    \edge {V2} {V1};
                    \edge {V3} {V2};
                    \edge {V5} {V4};
                    \edge {V2} {V5};
                    \edge {V2} {V4};
                    \node[coordinate] (C) at (0, 0.4) {};
                    \addvmargin{2mm};
                }
                \caption{Neighbouring DAG $\mathcal{G}_{n2}$}
                \label{fig: neighbouring dag 2}
            \end{subfigure}
        \end{minipage}
        \begin{tikzpicture}[remember picture, overlay]
            \draw[->] (A) -- (B) node[midway, above, rotate = 35]{add $V_6\rightarrow V_5$};
            \draw[->] (A) -- (C) node[midway, above, rotate = 325]{add $V_2\rightarrow V_4$};
        \end{tikzpicture}
    \end{minipage}
    \caption{Example of a searching step in HC-pairwise/HC-IPW.}
    \label{fig: example of a searching step in HC variants}
\end{figure}
Because the effectiveness of a scoring function increases with sample size, the scoring efficiency of HC-IPW can decrease considerably when missingness are MNAR for multiple variables. This is because both the number of partially observed variables and MNAR missingness increase the number of data cases removed during the learning process. It is on this basis we investigated a third variant, called the adaptive IPW-based HC (HC-aIPW), and which can be viewed as an extension of HC-IPW. The pseudo-code of HC-aIPW is shown in Algorithm~\ref{alg: hc-aipw algorithm}. The highlighted section represents the part of the code that differs from HC-IPW.

In essence, HC-aIPW aims to maximise the samples taken into consideration during the learning process. When there are partially observed variables in $\bm{Pa}_{\bm R_{\bm W}}$, HC-aIPW applies pairwise deletion given $\bm W$ and computes the difference in score between the current optimal DAG and the neighbouring DAG using the original pairwise deleted data set and standard scoring function. This is the only difference between HC-aIPW and HC-IPW. When there are no partially observed variables in $\bm{Pa}_{\bm R_{\bm W}}$, HC-aIPW uses the same IPW procedure as in HC-IPW to compute the difference in score between the current optimal DAG and the neighbouring DAG given the weighted pairwise deleted data set.
\begin{algorithm}[!ht]
    \caption{HC-aIPW algorithm}\label{alg: hc-aipw algorithm}
    \hspace*{\algorithmicindent} \textbf{Input} data set $D$\\
    \hspace*{\algorithmicindent} \textbf{Output} learned DAG $\mathcal{G}$
    \begin{algorithmic}[1]
        \Procedure{HC-aIPW}{}
        \State $\mathcal{G}\leftarrow$ empty graph
        \State $\mathcal{G}_{record}\leftarrow\left\{\mathcal{G}\right\}$
        \State retrieve the parents of missing indicators via Algorithm~\ref{alg: direct cause detection}
        \Repeat
            \State $\delta\leftarrow 0$
            \Repeat
                \State\parbox[t]{\dimexpr0.9\linewidth-\algorithmicindent}{construct a neighbouring DAG $\mathcal{G}_{nei}$ by \emph{adding}, \emph{reversing} or \emph{deleting} an edge from $\mathcal{G}$}
                \If{$\mathcal{G}_{nei}\not\in\mathcal{G}_{record}$}
                    \tikzmark{begin}\If{$\bm{Pa}_{\bm R_{\bm W}}\cap\bm V_m\neq\text{\O}$}
                        \State construct $D_{pw}$ by pairwise deleting $D$ given the necessary variables $\bm W$
                        \If{$S(\mathcal{G}_{nei}\mid D_{pw}) - S(\mathcal{G}\mid D_{pw}) > \delta$}
                            \State $\delta\leftarrow S(\mathcal{G}_{nei}\mid D_{pw}) - S(\mathcal{G}\mid D_{pw})$
                            \State $\mathcal{G}_{update}\leftarrow\mathcal{G}_{nei}$
                        \EndIf\tikzmark{end}
                    \Else
                        \State construct $D_{pw}$ by pairwise deleting $D$ given the sufficient variables $\bm U$
                        \State compute weight $\beta$ by Equation~\ref{equ: recover function} for $D_{pw}$
                        \If{$S(\mathcal{G}_{nei}\mid D_{pw}, \beta) - S(\mathcal{G}\mid D_{pw}, \beta) > \delta$}
                            \State $\delta\leftarrow S(\mathcal{G}_{nei}\mid D_{pw}, \beta) - S(\mathcal{G}\mid D_{pw}, \beta)$
                            \State $\mathcal{G}_{update}\leftarrow\mathcal{G}_{nei}$
                        \EndIf
                    \EndIf
                \EndIf
            \Until{all possible edge operations have been attempted}
            \If{$\delta > 0$}
                \State $\mathcal{G}\leftarrow\mathcal{G}_{update}$
                \State $\mathcal{G}_{record}=\mathcal{G}_{record}\cup\left\{\mathcal{G}\right\}$
            \EndIf
        \Until{$\delta=0$}
        \EndProcedure
    \end{algorithmic}\drawredbox
\end{algorithm}
\section{Experiments}
\label{sec: experiments}
The learning accuracy of each of the three algorithms described in Section~\ref{sec: handling missing data in hill-climbing algorithm} is investigated and evaluated with reference to the Structural EM algorithm when applied to the same data. The Structural EM algorithm represents a state-of-the-art score-based approach for structure learning from missing data, and also explores the search space of graphs using HC. Since all the involved algorithms are based on HC, we measure their learning accuracy with reference to the results obtained when applying standard HC on complete, rather than incomplete, data. Results from complete data give us the empirical maximum performance we can achieve on these data sets using HC, before making part of the data missing. The HC and Structural EM algorithms used in this paper are those available in the \textit{bnlearn} R package~\citep{scutari2010learning}. It is worth noting that the Structural EM algorithm implemented in \textit{bnlearn} R package is based on single imputation rather than belief propagation. Therefore, the results presented in this paper approximate the difference between the proposed methods and Friedman's Structural EM. The implementations of the three HC variants described in Section~\ref{sec: handling missing data in hill-climbing algorithm} are available online at~\url{https://github.com/Enderlogic/HC-missing-data}.
\subsection{Generating synthetic data and missingness}
\label{subsec: data description}
To illustrate the performance of the algorithms under different settings, we consider three types of ground truth DAGs: sparse networks, dense networks and real-world networks. We have constructed 50 random sparse and 50 random dense DAGs. Each network contains 20 to 50 nodes with two to six states per node. A sparse DAG $\mathcal{G}$ with $n$ variables is generated from a randomly ordered variable set $V_1<V_2<\ldots<V_n$, where directed edges are sampled from lower ordered variables to higher ordered variables with probability $2/\left(n - 1\right)$. Dense DAGs are generated with the same procedure, but the probability of drawing an edge between variables increases to $4/\left(n - 1\right)$. The conditional probability distribution of variable $V_i$ in sparse and dense DAGs is parameterised, given any configuration of its parents, by drawing a random number from the Dirichlet distribution $\text{Dir}\left(\bm\alpha\right)$, where $\bm\alpha = \underbrace{\left\{1, \ldots, 1\right\}}_{r_i}$, and $r_i$ is the number of states in $V_i$. For real-world DAGs, we use the six real-world BNs investigated in~\citep{constantinou2021large}. The structure and parameters of these BNs are set by either real data observations or prior knowledge as defined in the original studies. The properties of these BNs are provided in Table~\ref{tab: summary of real-world bns}.
\begin{table}[!ht]
\centering
\caption{The properties of the six real-world BNs.}
\begin{tabular}{cccc}
    \toprule
     Name & Number of variables & Average degree & Number of states\\
     \midrule
     Asia & 8 & 2.00 & 2\\
     Alarm & 37 & 2.49 & $2\sim 4$\\
     Pathfinder & 109 & 3.58 & $2\sim 63$\\
     Sports & 9 & 3.33 & $3\sim 8$\\
     ForMed & 88 & 3.14 & $2\sim 10$\\
     Property & 27 & 2.30 & $2\sim 7$\\
     \bottomrule
\end{tabular}
\label{tab: summary of real-world bns}
\end{table}%

We generate complete and incomplete synthetic data using the DAGs introduced above. The complete data sets are provided as input to the standard HC algorithm, whereas the corresponding incomplete data sets are provided as input to the Structural EM and the three HC variants described in Section~\ref{sec: handling missing data in hill-climbing algorithm}. We generate five complete data sets per DAG with sample sizes $N\in\left\{100, 500, 1000, 5000, 10000\right\}$. Each complete data set is then used to construct further three data sets with missing values; one per missingness assumption, MCAR, MAR or MNAR. For the MCAR case, we randomly select 50\% of the variables to represent the partially observed variables, and we then remove observed data of these variables with probability p, where p represents a random value between 0.1 and 0.6. For case MAR, we had to ensure missingness are dependent on a subset of the fully observed variables, and this is done as follows:
\begin{enumerate}
    \item Randomly select 50\% of the variables as partially observed variables (same process as in MCAR);
    \item Randomly assign a fully observed variable as the parent of missingness of a partially observed variable (repeat for all partially observed variables);
    \item Remove observations in partially observed variables with probability $p=0.6$ when the parent of their missingness is at its highest occurring state; otherwise, remove the observation with probability $p=0.1$.
\end{enumerate}
Generating MNAR data also involves the above 3-step procedure, but step 2 is modified as follows:
\begin{enumerate}
  \setcounter{enumi}{1}
  \item Randomly select 50\% of the partially observed variables and randomly assign a fully observed variable as the parent of their missingness. For the remaining 50\% partially observed variables, randomly assign another partially observed variable as the parent of their missingness.
\end{enumerate}
\subsection{Evaluation metrics}
\label{subsec: evaluation metrics}
The structure learning performance is assessed using two metrics that are fully oriented towards graphical discovery. The first metric is the classic $F_1$ score, composed of \emph{Precision} and \emph{Recall}. The formal definition of the $F_1$ score is:
\begin{equation}
    F_1 = 2\frac{\text{precision}*\text{recall}}{\text{precision}+\text{recall}} = \frac{2\,TP}{2\,TP+FP+FN}
\end{equation}
where $TP$ is the number of edges that exist in both the learned graph and true graph, $FP$ is the number of edges that exist in the learned graph but not in true graph, and $FN$ is the number of edges that exist in the true graph but not in the learned graph.

The second metric considered is the Structural Hamming Distance (SHD) which measures graphical differences between the learned graph and the true graph~\citep{tsamardinos2006max}. Specifically, the SHD score represents the number of edge operations needed to convert the learned graph to the true graph, where the edge operations involve arc addition, deletion and removal. Therefore, in contrast to the $F_1$ score, a lower SHD score indicates a better performance. Because the SHD score is sensitive to the number of edges and variables present in the true graph, we divide the SHD score by the number of edges in the true DAG to reduce bias.

Because the experiments are based on observational data, multiple DAGs can be statistically indistinguishable due to being part of the same Markov Equivalence class. On this basis, we compare the CPDAGs between the learned and true graphs to measure both the $F_1$ and SHD graphical scores.
\subsection{Results when the true DAG is sparse}
\label{subsec: results under sparse true dags}
\begin{figure}[!ht]
    \centering
    \includegraphics[width=\textwidth]{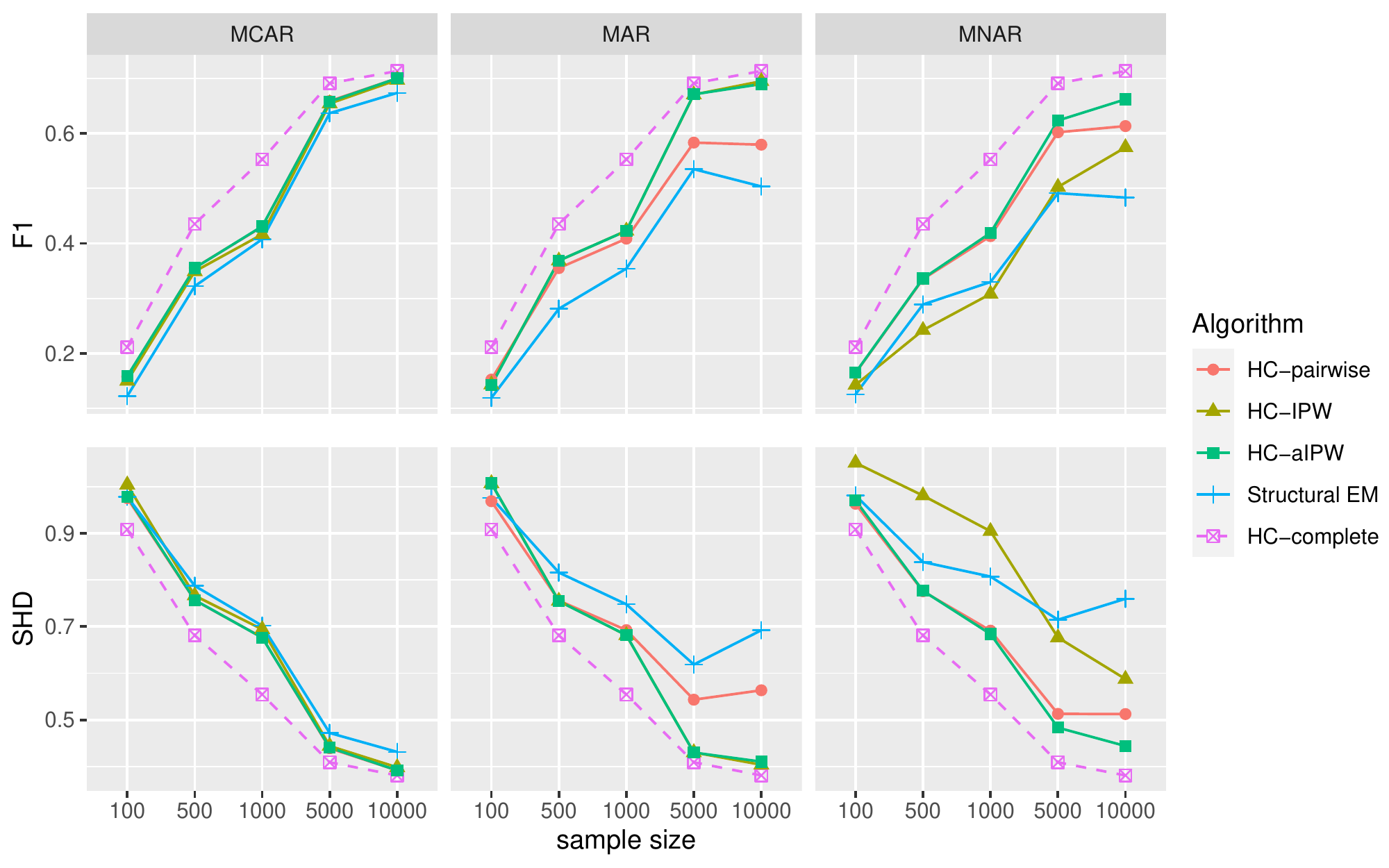}
    \caption{Average $F_1$ and normalised SHD scores learned by HC-pairwise, HC-IPW, HC-aIPW and Structural EM for sparse networks, under different assumptions of missingness and sample sizes. Each score represents the average score over 50 CPDAGs. Note the scores of HC-complete are based on complete data for benchmarking purposes; i.e., the same scores are superimposed in all three missingness cases as a dashed line.}
    \label{fig: f1 and shd under sparse true dags}
\end{figure}

Figure~\ref{fig: f1 and shd under sparse true dags} presents the average accuracy of the algorithms when the true DAGs are sparse. Each averaged score is derived from 50 CPDAGs, corresponding to each of the 50 randomly generated sparse DAGs. Appendix~\ref{app: accuracy results} provides the mean and standard deviation of the scores. The results suggest that the two evaluation metrics are generally consistent in ranking the algorithms from best to worst performance. Both metrics suggest that all of the three proposed HC variants outperform the Structural EM algorithm when the sample size is greater than 1,000, under all three missingness scenarios MCAR, MAR and MNAR. Interestingly, the HC-aIPW algorithm almost matches the performance of HC which is applied to complete data (denoted as HC-complete in Figures~\ref{fig: f1 and shd under sparse true dags}), particularly for experiments with 10,000 sample size, and this observation is consistent across all three missingness assumptions.

The three variants, HC-pairwise, HC-IPW and HC-aIPW, produce very similar results under MCAR, and this is because missingness under MCAR has no pattern that could be identified by the HC-IPW and HC-aIPW variants. That is, when HC-IPW and HC-aIPW do not discover any parent of missingness, they follow the search process of HC-pairwise. Under MAR, however, both HC-IPW and HC-aIPW outperform HC-pairwise as well as Structural EM when the sample size is larger than 100 and the improvement in performance increases with sample size. From this observation, we can conclude that the IPW method successfully eliminate most of the distributional bias. Interestingly, although the construction of the Structural EM algorithm is based on the MAR assumption, its performance under MAR is considerably lower than its performance under MCAR. A possible explanation is that the single imputation process the \emph{bnlearn} R package employs during the E step of Structural EM, instead of belief propagation, is unable to capture the uncertainty of the missing values.

Lastly, the results under MNAR suggest that HC-IPW generally performs worse than HC-pairwise across most sample sizes. This observation can be explained by the reduced sample size on which HC-IPW operates, relative to HC-pairwise, as discussed in subsection~\ref{subsec: hill-climbing with adaptive inverse probability weighting}. Specifically, when the parents of missingness of necessary variables $W$ contain partially observed variables (i.e., MNAR case), HC-IPW applies pairwise deletion by taking into consideration a higher number of variables compared to those considered by HC-pairwise. This means that, compared to HC-pairwise, the HC-IPW algorithm typically evaluates edge operations based on smaller samples when missingness are MNAR, which tends to yield less accurate results. From this, we can also conclude that the negative effect resulting from HC-IPW further pruning samples has not been offset by the data bias adjustments applied by the IPW method. On the other hand, the HC-aIPW algorithm which is designed to apply the IPW method only when no additional samples would be deleted compared to HC-pairwise, generally outperforms all other algorithms under MNAR, particularly under higher sample sizes.

\begin{figure}[!ht]
    \centering
    \includegraphics[width=\textwidth]{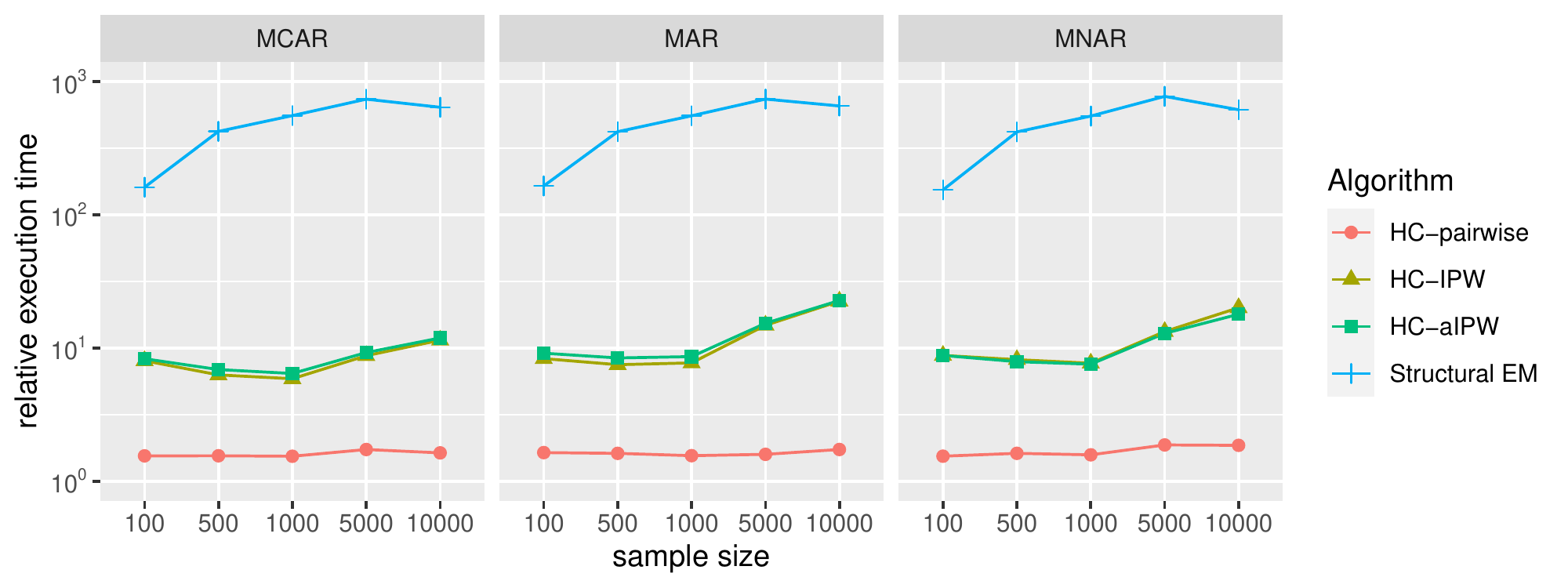}
    \caption{Average ratio of the execution time between the algorithms running on missing data sets and HC running on complete data sets.}
    \label{fig:time}
\end{figure}

Figure~\ref{fig:time} presents the relative execution time between a) the four algorithms applied to data with missing values, and b) the HC algorithm applied to the complete data. Because the three HC variants are implemented in Python, we measure their execution time relative to our Python version of HC. On the other hand, Structural EM is implemented in \emph{bnlearn} R package and makes use of the HC implementation of that package. Therefore, the execution time of Structural EM is measured relative to the HC implementation in \emph{bnlearn} R package. The mean and standard deviation of the results can be found in Appendix~\ref{app: execution time results}.

Overall, the results show that HC-pairwise is the most efficient algorithm for missingness. Specifically, HC-pairwise increases execution time relative to HC by approximately 50\%, while HC-IPW and HC-aIPW are anywhere between 8 and 15 times slower than HC dependent on sample size, and the relative difference in execution time tends to increase with sample size. This is because a higher number of parents of missingness are likely to be detected in larger sample sizes, and these discoveries increase execution time for IPW-based variants. Still, both the HC-IPW and HC-aIPW variants are more efficient than Structural EM which increases execution time relative to HC by 100 to 700 times.
\subsection{Results when the true DAG is dense}
\label{subsec: results under dense true dags}
In this subsection we investigate the performance of the algorithms when applied to data sets sampled from dense networks. The performance of each algorithm is depicted in Figure~\ref{fig: f1 and shd under dense true dags}, and detailed results are provided in Appendix~\ref{app: accuracy results}. An important distinction between sparse and dense networks is that learning from data sampled from dense networks makes it more likely that local parts of the graph will involve learning from partially observed variables. In other words, the effect of missing values is more severe on dense, compared to sparse, networks as shown in subsection~\ref{subsec: results under sparse true dags}.
\begin{figure}[!ht]
    \centering
    \includegraphics[width=\textwidth]{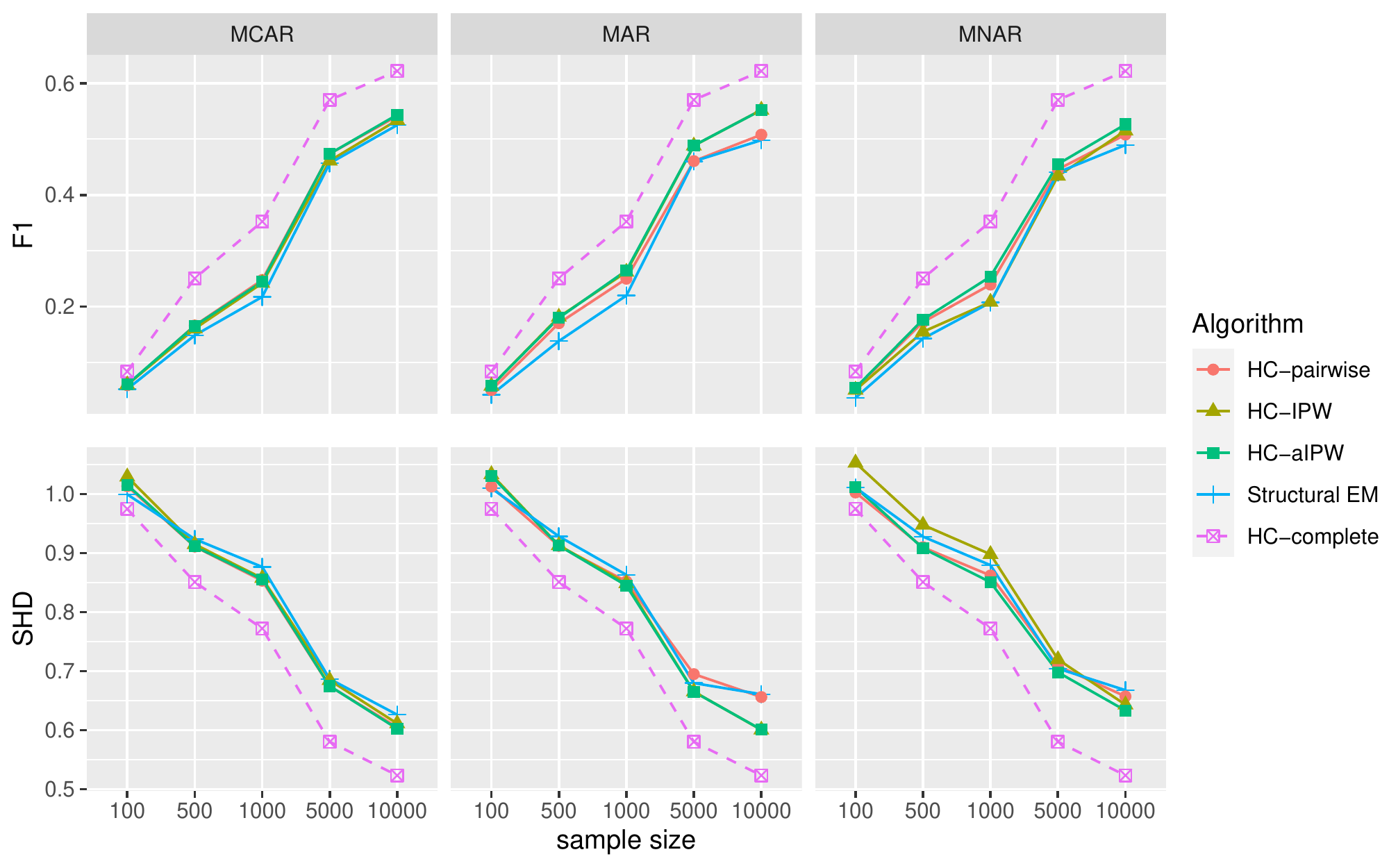}
    \caption{Average $F_1$ and normalised SHD scores learned by HC-pairwise, HC-IPW, HC-aIPW and Structural EM for dense networks, under different assumptions of missingness and sample sizes. Each score represents the average score over 50 CPDAGs. Note the scores of HC-complete are based on complete data for benchmarking purposes; i.e., the same scores are superimposed in all three missingness cases as a dashed line.}
    \label{fig: f1 and shd under dense true dags}
\end{figure}

The results show that the HC-aIPW algorithm continues to perform best in the case of denser graphs, in terms of overall performance and over the different missingness and sample size assumptions. Specifically, HC-aIPW achieves the highest accuracy in 11 and 8 cases in terms of $F_1$ and SHD measures respectively, out of the 15 experiments conducted in this subsection. In contrast, the Structural EM algorithm performs best only in two experiments and only in SHD score. However, compared with the results in subsection~\ref{subsec: results under sparse true dags}, the divergence in score between Structural EM and HC-based variants is much smaller.

The performance across the three HC-based variants appears to be similar to that obtained under sparse graphs. When data are MCAR, HC-IPW and HC-aIPW produce scores that are similar to those produced by HC-pairwise, and this is expected since no observed variables should be detected as the parents of missing indicators when missingness is MCAR. When data are MAR, both HC-IPW and HC-aIPW outperform HC-pairwise since, unlike HC-pairwise, they can detect and reduce bias caused by missing values. Lastly, when data are MNAR, HC-IPW performs worst amongst all algorithms, particularly when the sample size is lowest, and this is because it tends to remove a large number of data cases when computing the local scores. On the other hand, HC-aIPW (which aims to resolve this specific drawback of HC-IPW) performs best in almost all MNAR experiments. The consistency of the results across sparse and dense networks suggests that the performance of HC-aIPW, relative to the other algorithms considered in this study, is not sensitive to the sparsity of the network that generates the input data.
\subsection{Results when the true DAG is a real-world network}
\label{subsec: results under real-world true dags}
\begin{figure}[!ht]
    \centering
    \includegraphics[width=\textwidth]{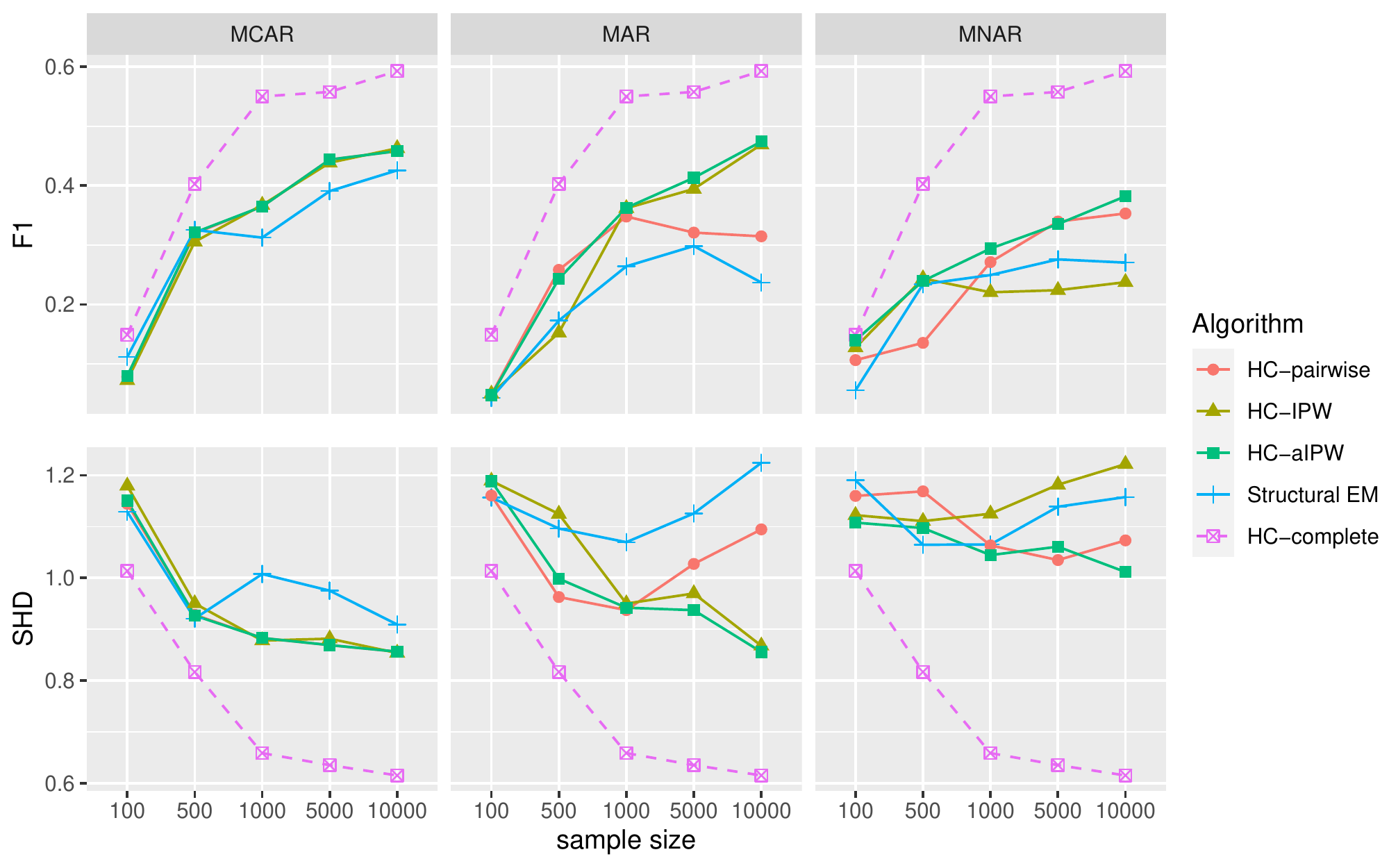}
    \caption{Average $F_1$ and normalised SHD scores learned by HC-pairwise, HC-IPW, HC-aIPW and Structural EM for real-world networks, under different assumptions of missingness and sample sizes. Each score represents the average score over 50 CPDAGs. Note the scores of HC-complete are based on complete data for benchmarking purposes; i.e., the same scores are superimposed in all three missingness cases as a dashed line.}
    \label{fig: f1 and shd under real-world true dags}
\end{figure}

Lastly, we apply the algorithms to data sets sampled from the six real-world networks. Figure~\ref{fig: f1 and shd under real-world true dags} shows the average performance of the algorithms across all the six real-world networks and over all the five sample sizes. When the missingness is MCAR, the three HC-based variants achieve similar accuracy, as expected, and generally outperform the Structural EM algorithm when the sample size is larger than 500. When the missingness is MAR or MNAR, the performance of HC-aIPW improves over the other algorithms, especially when the sample size is larger than 500. These results are consistent with those obtained from the randomised sparse and dense networks presented in subsections~\ref{subsec: results under sparse true dags} and~\ref{subsec: results under dense true dags} respectively.
\section{Conclusion}
\label{sec: conclusion}
Learning accurate BN structure from incomplete data remains a challenging task. Most BN structure learning algorithms do not support learning from incomplete data, and this is partly explained by the considerable increase in computational complexity when dealing with incomplete data. The increased computational complexity caused by missing data adds to a problem that is NP-hard even when data are complete. This challenge is even greater when missing values are systematic rather than random.

In this paper, we have investigated three novel HC-based variants that employ pairwise deletion and IPW strategies to deal with random and systematic missing data. The HC-pairwise and HC-IPW variants can be viewed as subversions of HC-aIPW, which is the most complete and best performing variant described in this paper. All of the three variants have been applied to different cases of data missingness, and their performance was compared to the state-of-the-art Structural EM algorithm that is available in the \emph{bnlearn} R package. Moreover, all performances under missingness have been compared to HC when applied to the corresponding complete data sets. The empirical results show:
\begin{enumerate}
    \item Pairing HC with pairwise deletion (i.e., the HC-pairwise variant) is enough to learn graphs that are more accurate, as well as less computationally expensive, compared to the graphs produced by the Structural EM algorithm.
    \item Combining HC with both pairwise deletion and IPW techniques (i.e., the HC-IPW variant) further improves learning accuracy under MCAR and MAR, in general, but decreases accuracy under MNAR due to aggressive pruning employed by HC-IPW on the data cases (refer to subsection~\ref{subsec: hill-climbing with adaptive inverse probability weighting}). Moreover, HC-IPW becomes considerably slower than HC-pairwise, although it remains an order of magnitude faster than Structural EM.
    \item The HC-aIPW takes advantage of both strategies, as in HC-IPW, but relaxes the pruning strategy on the data cases and returns the overall best performance, especially under MNAR which represents the most difficult case of missingness.
    \item All three HC variants described in this paper outperform Structural EM in most cases. Importantly, the performance of HC-aIPW on missing data approaches the performance of HC on complete data when sample size is 10,000 and the ground truth graph is sparse, and this observation is consistent under all three cases of missingness.
\end{enumerate}
Future research will investigate the application of these learning strategies to search algorithms that are more complex than HC, such as Tabu, or other variants of HC such as the GES algorithm~\citep{chickering2002optimal} which explores the CPDAG, rather than DAG space. Another possible research direction would be to combine the IPW method with the NAL score~\citep{balov2013consistent}, which is a scoring function intended for missingness under MCAR, and further investigate the possibility of a new decomposable scoring function under systematic missingness cases of MAR and MNAR.
\begin{appendices}

\section{Proofs of propositions}
\label{app: proofs of proposition}
In this section, we provide proofs of the propositions discussed in Section~\ref{sec: handling missing data in hill-climbing algorithm}. We define the variables used in proofs as follows: $\bm V_d$ is the set of variables with different parent-sets between a given DAG $\mathcal{G}$ and its neighbouring DAG $\mathcal{G}_{nei}$, $\bm W$ is a set of the necessary variables as defined in Equation~\ref{equ: necessary variables}, $\bm U$ is a set of the sufficient variables defined in Equation~\ref{equ: sufficient variables}, and $N$ and $N_{pw}$ are the sample sizes of the partially observed data set $D$ and pairwise deleted data set $D_{pw}$ respectively.
\mcar*
\allowdisplaybreaks
\begin{proof}
    \begin{align}
        S_{BIC}&\left(\mathcal{G}_{nei}\mid D_f\right) - S_{BIC}\left(\mathcal{G}\mid D_f\right)\nonumber\\ 
        & = \sum_{i = 1}^n\left(S_{BIC}(V_i\mid\bm{Pa}_i^{nei}) - S_{BIC}(V_i\mid\bm{Pa}_i)\right)\nonumber\\
        & = \sum_{i : V_i\in\bm V_d}\left(S_{BIC}(V_i\mid\bm{Pa}_i^{nei}) - S_{BIC}(V_i\mid\bm{Pa}_i)\right)\nonumber\\
        & = \sum_{i : V_i\in\bm V_d}\Bigg(\sum_{D_f}\left(\text{log}P(V_i\mid\bm{Pa}_i^{nei}) - \text{log}P(V_i\mid\bm{Pa}_i)\right)\nonumber\\
        &\qquad + \frac{\text{log}\left(N\right)}{2}\left(\lvert\hat{\Theta}_i^{nei}\rvert - \lvert\hat{\Theta}_i\rvert\right)\Bigg)\nonumber\\
        & = \frac{N}{N_{pw}}\sum_{i : V_i\in\bm V_d}\Bigg(\sum_{D_{pw}}\left( \text{log}P(V_i\mid\bm{Pa}_i^{nei}, \bm R_{\bm W} = \bm 0) - \text{log}P(V_i\mid\bm{Pa}_i, \bm R_{\bm W} = \bm 0)\right)\nonumber\\
        &\qquad + \frac{\text{log}\left(N_{pw}\right)}{2}\left(\lvert\hat{\Theta}_i^{nei}\rvert - \lvert\hat{\Theta}_i\rvert\right) + \frac{\text{log}\left(N/N_{pw}\right)}{2}\left(\lvert\hat{\Theta}_i^{nei}\rvert - \lvert\hat{\Theta}_i\rvert\right)\Bigg)\label{equ: relation to equation 5}\\
        & = \frac{N}{N_{pw}}\Bigg( S_{BIC}\left(\mathcal{G}_{nei}\mid D_{pw}\right) - S_{BIC}\left(\mathcal{G}\mid D_{pw}\right)\nonumber\\
        &\qquad + \frac{\text{log}\left(N/N_{pw}\right)}{2}\sum_{i : V_i\in\bm V_d}\left(\lvert\hat{\Theta}_i^{nei}\rvert - \lvert\hat{\Theta}_i\rvert\right)\Bigg)\nonumber\\
        &\propto S_{BIC}\left(\mathcal{G}_{nei}\mid D_{pw}\right) - S_{BIC}\left(\mathcal{G}\mid D_{pw}\right) + O(1)\label{equ: last equation in proof of proposition 1}
    \end{align}
    Equation~\ref{equ: relation to equation 5} follows from Equation~\ref{equ: local MCAR} given the MCAR assumption and large sample limit. Equation~\ref{equ: last equation in proof of proposition 1} is due to the missing rate of data $D$, i.e., $N_{pw} / N$, does not relate to the sample size $N$ and remains constant with the increase of $N$.
\end{proof}
\mnar*
\allowdisplaybreaks
\begin{proof}
    \begin{align}
        S_{BIC}&\left(\mathcal{G}_{nei}\mid D_f\right) - S_{BIC}\left(\mathcal{G}\mid D_f\right)\nonumber\\
        & = \sum_{i : V_i\in\bm V_d}\Bigg(\sum_{D_f}\left(\text{log}P(V_i\mid\bm{Pa}_i^{nei}) - \text{log}P(V_i\mid\bm{Pa}_i)\right)\nonumber\\
        &\qquad + \frac{\text{log}\left(N\right)}{2}\left(\lvert\hat{\Theta}_i^{nei}\rvert - \lvert\hat{\Theta}_i\rvert\right)\Bigg)\nonumber\\
        & = \sum_{i : V_i\in\bm V_d}\Bigg(\sum_{D_f}\left(\text{log}\frac{P(V_i, \bm{Pa}_i^{nei})}{\sum_{V_i}P(V_i,\bm{Pa}_i^{nei})} - \text{log}\frac{P(V_i, \bm{Pa}_i)}{\sum_{V_i}P(V_i, \bm{Pa}_i)}\right)\nonumber\\
        &\qquad + \frac{\text{log}\left(N\right)}{2}\left(\lvert\hat{\Theta}_i^{nei}\rvert - \lvert\hat{\Theta}_i\rvert\right)\Bigg)\nonumber\\
        & = \frac{N}{N_{pw}}\sum_{i : V_i\in\bm V_d}\Bigg(\sum_{D_{pw}}\Bigg(\text{log}\frac{P(V_i, \bm{Pa}_i^{nei}\mid\bm R_{\bm U} = \bm 0)\beta}{\sum_{V_i}P(V_i, \bm{Pa}_i^{nei}\mid\bm R_{\bm U} = \bm 0)\beta}\nonumber\\
        &\qquad - \text{log}\frac{P(V_i, \bm{Pa}_i\mid\bm R_{\bm U} = \bm 0)\beta}{\sum_{V_i}P(V_i, \bm{Pa}_i\mid\bm R_{\bm U}=\bm 0)\beta}\Bigg) + \frac{\text{log}\left(N\right)}{2}\left(\lvert\hat{\Theta}_i^{nei}\rvert - \lvert\hat{\Theta}_i\rvert\right)\Bigg)\label{equ: relation to equation 7}\\
        & = \frac{N}{N_{pw}}\sum_{i : V_i\in\bm V_d}\Bigg(\sum_{j = 1}^{\lvert\bm{Pa}_i^{nei}\rvert}\sum_{k = 1}^{\lvert V_i\rvert}\widetilde{N}_{ijk}\text{log}\frac{\widetilde{N}_{ijk}}{\widetilde{N}_{ij}} - \sum_{j = 1}^{\lvert\bm{Pa}_i\rvert}\sum_{k = 1}^{\lvert V_i\rvert}\widetilde{N}_{ijk}\text{log}\frac{\widetilde{N}_{ijk}}{\widetilde{N}_{ij}}\nonumber\\
        &\qquad + \frac{\text{log}\left(N\right)}{2}\left(\lvert\hat{\Theta}_i^{nei}\rvert - \lvert\hat{\Theta}_i\rvert\right)\Bigg)\nonumber\\
        & = \frac{N}{N_{pw}}\sum_{i : V_i\in\bm V_d}\Bigg(\sum_{j = 1}^{\lvert\bm{Pa}_i^{nei}\rvert}\sum_{k = 1}^{\lvert V_i\rvert}\widetilde{N}_{ijk}\text{log}\frac{\widetilde{N}_{ijk}}{\widetilde{N}_{ij}} - \sum_{j = 1}^{\lvert\bm{Pa}_i\rvert}\sum_{k = 1}^{\lvert V_i\rvert}\widetilde{N}_{ijk}\text{log}\frac{\widetilde{N}_{ijk}}{\widetilde{N}_{ij}}\nonumber\\
        &\qquad + \frac{\text{log}\left(N_{pw}\right)}{2}\left(\lvert\hat{\Theta}_i^{nei}\rvert - \lvert\hat{\Theta}_i\rvert\right) + \frac{\text{log}\left(N/N_{pw}\right)}{2}\left(\lvert\hat{\Theta}_i^{nei}\rvert - \lvert\hat{\Theta}_i\rvert\right)\Bigg)\nonumber\\
        & = \frac{N}{N_{pw}}\Bigg(S_{BIC}\left(\mathcal{G}_{nei}\mid D_{pw}, \beta\right) - S_{BIC}\left(\mathcal{G}\mid D_{pw}, \beta\right)\nonumber\\
        &\qquad + \frac{\text{log}\left(N/N_{pw}\right)}{2}\sum_{i : V_i\in\bm V_d}\left(\lvert\hat{\Theta}_i^{nei}\rvert - \lvert\hat{\Theta}_i\rvert\right)\Bigg)\nonumber\\
        &\propto S_{BIC}\left(\mathcal{G}_{nei}\mid D_{pw}, \beta\right) - S_{BIC}\left(\mathcal{G}\mid D_{pw}, \beta\right) + O(1)\nonumber
    \end{align}
    In the above equations, $\beta = \prod_{R_i\in\bm R_{\bm U}}\beta_{R_i}$, $\widetilde{N}_{ijk}$ and $\widetilde{N}_{ij}$ are defined by Equation~\ref{equ: nijk} and~\ref{equ: nij}. Equation~\ref{equ: relation to equation 7} is a consequence of the recoverability of $P\left(\bm U\right)$ given Equation~\ref{equ: recover function}.
\end{proof}
\section{Derivation of Equation~\ref{equ: recover function}}
\label{app: derivation of equation 7}
Based on~\cite{NIPS2013_0ff8033c}, Theorem 2, given Assumptions~\ref{ass: r is effect node} and~\ref{ass: no cause between variable and its own indicator}, the joint distribution $P\left(\bm V\right)$ can be fully recovered from the observed data via the following equation:
\begin{equation*}
    P\left(\bm V\right) = \frac{P\left(\bm V, \bm R = \bm 0\right)}{\prod_{R_i\in\bm R} P\left(R_i = 0\mid \bm{Pa}_{R_i}, \bm R_{\bm{Pa}_{R_i}} = \bm 0\right)}
\end{equation*}
where $\bm{Pa}_{R_i}$ is the set of parents of missing indicator $R_i$, and $\bm R_{\bm{Pa}_{R_i}}$ is the set of missing indicator of the partially observed variables in $\bm{Pa}_{R_i}$. Then,
\begin{align*}
    P\left(\bm V\right) &= \frac{P\left(\bm V, \bm R = \bm 0\right)}{\prod_{R_i\in\bm R} P\left(R_i = 0\mid \bm{Pa}_{R_i}, \bm R_{\bm{Pa}_{R_i}} = \bm 0\right)}\\
    &= \frac{P\left(\bm V\mid\bm R = \bm 0\right)P\left(\bm R = \bm 0\right)}{\prod_{R_i\in\bm R} P\left(R_i = 0\mid \bm{Pa}_{R_i}, \bm R_{\bm{Pa}_{R_i}} = \bm 0\right)}\\
    &= P\left(\bm V\mid\bm R = \bm 0\right)\cdot\frac{P\left(\bm R = \bm 0\right)}{\prod_{R_i\in\bm R}\frac{P\left(\bm{Pa}_{R_i}\mid R_i = 0, \bm R_{\bm{Pa}_{R_i}} = \bm 0\right)P\left(R_i = 0\mid \bm R_{\bm{Pa}_{R_i}} = \bm 0\right)}{P\left(\bm{Pa}_{R_i}\mid \bm R_{\bm{Pa}_{R_i}} = \bm 0\right)}}\\
    &= P\left(\bm V\mid\bm R = \bm 0\right)\cdot\\
    &\qquad\underbrace{\frac{P\left(\bm R = \bm 0\right)}{\prod_{R_i\in\bm R}P\left(R_i = 0\mid \bm R_{\bm{Pa}_{R_i}} = \bm 0\right)}}_c\prod_{R_i\in\bm R}\underbrace{\frac{P\left(\bm{Pa}_{R_i}\mid \bm R_{\bm{Pa}_{R_i}} = \bm 0\right)}{P\left(\bm{Pa}_{R_i}\mid R_i = 0, \bm R_{\bm{Pa}_{R_i}} = \bm 0\right)}}_{\beta_{R_i}}
\end{align*}
In the above equation, the term $c$ depends only on the missing indicators $\bm R$ and remains constant with respect to the observed variables $\bm V$. The product $\prod_{R_i\in\bm R}\beta_{R_i}$ represents the relative probability of a data case from the pairwise deleted data set being observed in the complete data set. For example, if a pairwise deleted data case has $c\prod_{R_i\in \bm R}\beta_{R_i}$ out of 0.8, then its occurrence rate is assumed to drop by 20\% in the complete data set compared to its occurrence rate in the pairwise deleted data set. Therefore, we use Equation~\ref{equ: recover function} to reweight the pairwise deleted data and estimate the underlying true distribution given the pairwise deleted data set.
\section{Supplementary results from the structure learning experiments}
\label{app: accuracy results}
Refer to Table~\ref{tab: f1 on sparse true dags}, ~\ref{tab: shd on sparse true dags}, ~\ref{tab: f1 on dense true dags}, ~\ref{tab: shd on dense true dags}, ~\ref{tab: f1 on real-world true dags} and~\ref{tab: shd on real-world true dags}.
\begin{table}[H]
\centering
\caption{Mean and standard deviation of $F_1$ scores produced by Structural EM, HC-pairwise, HC-IPW and HC-aIPW for sparse networks, under the different assumptions of missingness and sample sizes.}
\begin{tabular}{>{\centering\arraybackslash}m{1cm}>{\centering\arraybackslash}m{1cm}>{\centering\arraybackslash}m{1.9cm}>{\centering\arraybackslash}m{1.9cm}>{\centering\arraybackslash}m{1.9cm}>{\centering\arraybackslash}m{1.9cm}}
    \toprule
     data & sample size & Structural EM & HC-pairwise & HC-IPW & HC-aIPW \\
     \midrule\multirow{5}{*}{MCAR} & 100 & $0.122\pm0.088$ & $0.158\pm0.108$ & $0.150\pm0.104$ & $\bm{0.159\pm0.107}$\\ & 500 & $0.325\pm0.139$ & $\bm{0.356\pm0.139}$ & $0.349\pm0.133$ & $0.355\pm0.139$\\ & 1000 & $0.410\pm0.141$ & $0.430\pm0.149$ & $0.417\pm0.143$ & $\bm{0.431\pm0.150}$\\ & 5000 & $0.642\pm0.149$ & $\bm{0.659\pm0.144}$ & $0.654\pm0.160$ & $0.658\pm0.144$\\ & 10000 & $0.682\pm0.135$ & $0.700\pm0.140$ & $0.697\pm0.143$ & $\bm{0.700\pm0.137}$\\\midrule\multirow{5}{*}{MAR} & 100 & $0.117\pm0.097$ & $\bm{0.152\pm0.102}$ & $0.143\pm0.102$ & $0.142\pm0.101$\\ & 500 & $0.281\pm0.119$ & $0.355\pm0.117$ & $0.369\pm0.140$ & $\bm{0.369\pm0.138}$\\ & 1000 & $0.354\pm0.136$ & $0.409\pm0.152$ & $0.423\pm0.150$ & $\bm{0.423\pm0.149}$\\ & 5000 & $0.543\pm0.119$ & $0.583\pm0.137$ & $\bm{0.671\pm0.147}$ & $0.671\pm0.150$\\ & 10000 & $0.505\pm0.141$ & $0.580\pm0.121$ & $\bm{0.695\pm0.136}$ & $0.690\pm0.138$\\\midrule\multirow{5}{*}{MNAR} & 100 & $0.127\pm0.094$ & $0.164\pm0.098$ & $0.143\pm0.103$ & $\bm{0.165\pm0.099}$\\ & 500 & $0.285\pm0.122$ & $0.335\pm0.129$ & $0.242\pm0.091$ & $\bm{0.336\pm0.131}$\\ & 1000 & $0.328\pm0.123$ & $0.413\pm0.142$ & $0.308\pm0.113$ & $\bm{0.419\pm0.148}$\\ & 5000 & $0.488\pm0.137$ & $0.602\pm0.164$ & $0.503\pm0.124$ & $\bm{0.624\pm0.150}$\\ & 10000 & $0.473\pm0.148$ & $0.613\pm0.144$ & $0.575\pm0.157$ & $\bm{0.662\pm0.146}$\\
     \bottomrule
\end{tabular}
\label{tab: f1 on sparse true dags}
\end{table}%
\begin{table}[H]
\centering
\caption{Mean and standard deviation of normalised SHD scores produced by Structural EM, HC-pairwise, HC-IPW and HC-aIPW for sparse networks, under the different assumptions of missingness and sample sizes.}
\begin{tabular}{>{\centering\arraybackslash}m{1cm}>{\centering\arraybackslash}m{1cm}>{\centering\arraybackslash}m{1.9cm}>{\centering\arraybackslash}m{1.9cm}>{\centering\arraybackslash}m{1.9cm}>{\centering\arraybackslash}m{1.9cm}}
    \toprule
     data & sample size & Structural EM & HC-pairwise & HC-IPW & HC-aIPW \\
     \midrule\multirow{5}{*}{MCAR} & 100 & $0.978\pm0.063$ & $\bm{0.975\pm0.095}$ & $1.004\pm0.107$ & $0.978\pm0.093$\\ & 500 & $0.784\pm0.121$ & $\bm{0.756\pm0.127}$ & $0.766\pm0.120$ & $0.757\pm0.127$\\ & 1000 & $0.700\pm0.133$ & $0.677\pm0.147$ & $0.694\pm0.140$ & $\bm{0.676\pm0.147}$\\ & 5000 & $0.465\pm0.181$ & $\bm{0.439\pm0.178}$ & $0.444\pm0.191$ & $0.441\pm0.179$\\ & 10000 & $0.424\pm0.178$ & $0.392\pm0.182$ & $0.398\pm0.183$ & $\bm{0.392\pm0.178}$\\\midrule\multirow{5}{*}{MAR} & 100 & $0.975\pm0.086$ & $\bm{0.969\pm0.094}$ & $1.007\pm0.101$ & $1.007\pm0.101$\\ & 500 & $0.814\pm0.103$ & $0.756\pm0.108$ & $0.755\pm0.131$ & $\bm{0.754\pm0.129}$\\ & 1000 & $0.748\pm0.123$ & $0.692\pm0.160$ & $\bm{0.681\pm0.156}$ & $0.682\pm0.154$\\ & 5000 & $0.609\pm0.158$ & $0.543\pm0.174$ & $\bm{0.430\pm0.186}$ & $0.430\pm0.189$\\ & 10000 & $0.690\pm0.188$ & $0.563\pm0.161$ & $\bm{0.404\pm0.181}$ & $0.411\pm0.184$\\\midrule\multirow{5}{*}{MNAR} & 100 & $0.984\pm0.068$ & $\bm{0.963\pm0.078}$ & $1.051\pm0.129$ & $0.970\pm0.082$\\ & 500 & $0.836\pm0.102$ & $\bm{0.776\pm0.114}$ & $0.980\pm0.116$ & $0.777\pm0.115$\\ & 1000 & $0.810\pm0.119$ & $0.691\pm0.140$ & $0.904\pm0.163$ & $\bm{0.684\pm0.149}$\\ & 5000 & $0.721\pm0.184$ & $0.513\pm0.201$ & $0.676\pm0.177$ & $\bm{0.483\pm0.185}$\\ & 10000 & $0.774\pm0.201$ & $0.513\pm0.185$ & $0.588\pm0.203$ & $\bm{0.444\pm0.184}$\\
     \bottomrule
\end{tabular}
\label{tab: shd on sparse true dags}
\end{table}%
\begin{table}[H]
\centering
\caption{Mean and standard deviation of $F_1$ scores produced by Structural EM, HC-pairwise, HC-IPW and HC-aIPW for dense networks, under the different assumptions of missingness and sample sizes.}
\begin{tabular}{>{\centering\arraybackslash}m{1cm}>{\centering\arraybackslash}m{1cm}>{\centering\arraybackslash}m{1.9cm}>{\centering\arraybackslash}m{1.9cm}>{\centering\arraybackslash}m{1.9cm}>{\centering\arraybackslash}m{1.9cm}}
    \toprule
     data & sample size & Structural EM & HC-pairwise & HC-IPW & HC-aIPW \\
     \midrule\multirow{5}{*}{MCAR} & 100 & $0.052\pm0.043$ & $0.059\pm0.046$ & $\bm{0.060\pm0.045}$ & $0.060\pm0.046$\\ & 500 & $0.148\pm0.070$ & $0.166\pm0.077$ & $0.160\pm0.071$ & $\bm{0.166\pm0.076}$\\ & 1000 & $0.217\pm0.094$ & $\bm{0.248\pm0.089}$ & $0.242\pm0.088$ & $0.245\pm0.092$\\ & 5000 & $0.457\pm0.136$ & $\bm{0.474\pm0.124}$ & $0.461\pm0.115$ & $\bm{0.474\pm0.124}$\\ & 10000 & $0.525\pm0.140$ & $0.541\pm0.151$ & $0.534\pm0.147$ & $\bm{0.544\pm0.148}$\\\midrule\multirow{5}{*}{MAR} & 100 & $0.042\pm0.048$ & $0.050\pm0.052$ & $0.058\pm0.053$ & $\bm{0.058\pm0.051}$\\ & 500 & $0.138\pm0.071$ & $0.170\pm0.081$ & $\bm{0.181\pm0.081}$ & $0.180\pm0.078$\\ & 1000 & $0.220\pm0.109$ & $0.250\pm0.095$ & $0.262\pm0.086$ & $\bm{0.265\pm0.088}$\\ & 5000 & $0.460\pm0.124$ & $0.461\pm0.121$ & $0.488\pm0.129$ & $\bm{0.488\pm0.128}$\\ & 10000 & $0.498\pm0.118$ & $0.508\pm0.126$ & $\bm{0.552\pm0.132}$ & $0.552\pm0.133$\\\midrule\multirow{5}{*}{MNAR} & 100 & $0.036\pm0.040$ & $0.054\pm0.053$ & $0.051\pm0.054$ & $\bm{0.054\pm0.052}$\\ & 500 & $0.143\pm0.066$ & $0.172\pm0.083$ & $0.155\pm0.064$ & $\bm{0.177\pm0.084}$\\ & 1000 & $0.207\pm0.087$ & $0.239\pm0.093$ & $0.208\pm0.077$ & $\bm{0.254\pm0.100}$\\ & 5000 & $0.440\pm0.123$ & $0.446\pm0.118$ & $0.434\pm0.126$ & $\bm{0.455\pm0.124}$\\ & 10000 & $0.490\pm0.125$ & $0.508\pm0.115$ & $0.515\pm0.127$ & $\bm{0.526\pm0.127}$\\
     \bottomrule
\end{tabular}
\label{tab: f1 on dense true dags}
\end{table}%
\begin{table}[H]
\centering
\caption{Mean and standard deviation of normalised SHD scores produced by Structural EM, HC-pairwise, HC-IPW and HC-aIPW for dense networks, under the different assumptions of missingness and sample sizes.}
\begin{tabular}{>{\centering\arraybackslash}m{1cm}>{\centering\arraybackslash}m{1cm}>{\centering\arraybackslash}m{1.9cm}>{\centering\arraybackslash}m{1.9cm}>{\centering\arraybackslash}m{1.9cm}>{\centering\arraybackslash}m{1.9cm}}
    \toprule
     data & sample size & Structural EM & HC-pairwise & HC-IPW & HC-aIPW \\
     \midrule
     \multirow{5}{*}{MCAR} & 100 & $\bm{0.999\pm0.033}$ & $1.013\pm0.036$ & $1.029\pm0.046$ & $1.015\pm0.037$\\
     & 500 & $0.924\pm0.047$ & $\bm{0.911\pm0.053}$ & $0.915\pm0.050$ & $\bm{0.911\pm0.053}$\\
     & 1000 & $0.876\pm0.068$ & $\bm{0.853\pm0.072}$ & $0.857\pm0.071$ & $0.855\pm0.074$\\
     & 5000 & $0.687\pm0.137$ & $\bm{0.674\pm0.131}$ & $0.685\pm0.118$ & $0.675\pm0.131$\\
     & 10000 & $0.626\pm0.156$ & $0.605\pm0.175$ & $0.611\pm0.169$ & $\bm{0.603\pm0.173}$\\
     \midrule
     \multirow{5}{*}{MAR} & 100 & $\bm{1.010\pm0.032}$ & $1.013\pm0.031$ & $1.033\pm0.050$ & $1.031\pm0.045$\\
     & 500 & $0.928\pm0.049$ & $\bm{0.912\pm0.056}$ & $0.913\pm0.062$ & $0.913\pm0.062$\\
     & 1000 & $0.863\pm0.089$ & $0.852\pm0.076$ & $0.848\pm0.067$ & $\bm{0.845\pm0.070}$\\
     & 5000 & $0.680\pm0.133$ & $0.695\pm0.133$ & $0.665\pm0.141$ & $\bm{0.665\pm0.140}$\\
     & 10000 & $0.661\pm0.144$ & $0.656\pm0.154$ & $\bm{0.601\pm0.157}$ & $0.601\pm0.158$\\
     \midrule
     \multirow{5}{*}{MNAR} & 100 & $1.011\pm0.035$ & $\bm{1.003\pm0.037}$ & $1.053\pm0.064$ & $1.011\pm0.041$\\
     & 500 & $0.928\pm0.044$ & $0.910\pm0.052$ & $0.948\pm0.052$ & $\bm{0.908\pm0.053}$\\
     & 1000 & $0.879\pm0.064$ & $0.862\pm0.073$ & $0.898\pm0.062$ & $\bm{0.851\pm0.079}$\\
     & 5000 & $0.704\pm0.129$ & $0.709\pm0.119$ & $0.720\pm0.134$ & $\bm{0.698\pm0.127}$\\
     & 10000 & $0.668\pm0.142$ & $0.657\pm0.135$ & $0.643\pm0.144$ & $\bm{0.633\pm0.151}$\\
     \bottomrule
\end{tabular}
\label{tab: shd on dense true dags}
\end{table}%
\begin{table}[H]
\centering
\caption{Mean and standard deviation of $F_1$ scores produced by Structural EM, HC-pairwise, HC-IPW and HC-aIPW for real-world networks, under the different assumptions of missingness and sample sizes.}
\begin{tabular}{>{\centering\arraybackslash}m{1cm}>{\centering\arraybackslash}m{1cm}>{\centering\arraybackslash}m{1.9cm}>{\centering\arraybackslash}m{1.9cm}>{\centering\arraybackslash}m{1.9cm}>{\centering\arraybackslash}m{1.9cm}}
    \toprule
     data & sample size & Structural EM & HC-pairwise & HC-IPW & HC-aIPW \\
     \midrule\multirow{5}{*}{MCAR} & 100 & $\bm{0.112\pm0.103}$ & $0.078\pm0.048$ & $0.072\pm0.046$ & $0.079\pm0.050$\\ & 500 & $\bm{0.325\pm0.268}$ & $0.319\pm0.273$ & $0.305\pm0.279$ & $0.321\pm0.272$\\ & 1000 & $0.312\pm0.265$ & $0.365\pm0.220$ & $\bm{0.367\pm0.216}$ & $0.365\pm0.220$\\ & 5000 & $0.391\pm0.248$ & $\bm{0.444\pm0.213}$ & $0.438\pm0.219$ & $\bm{0.444\pm0.213}$\\ & 10000 & $0.426\pm0.246$ & $0.458\pm0.196$ & $\bm{0.463\pm0.192}$ & $0.458\pm0.196$\\\midrule\multirow{5}{*}{MAR} & 100 & $0.043\pm0.037$ & $0.048\pm0.039$ & $\bm{0.050\pm0.041}$ & $0.047\pm0.039$\\ & 500 & $0.173\pm0.149$ & $\bm{0.258\pm0.338}$ & $0.152\pm0.121$ & $0.243\pm0.342$\\ & 1000 & $0.264\pm0.303$ & $0.348\pm0.352$ & $0.362\pm0.313$ & $\bm{0.363\pm0.311}$\\ & 5000 & $0.298\pm0.340$ & $0.321\pm0.256$ & $0.394\pm0.266$ & $\bm{0.413\pm0.261}$\\ & 10000 & $0.237\pm0.326$ & $0.315\pm0.247$ & $0.469\pm0.349$ & $\bm{0.474\pm0.343}$\\\midrule\multirow{5}{*}{MNAR} & 100 & $0.055\pm0.047$ & $0.106\pm0.095$ & $0.127\pm0.153$ & $\bm{0.140\pm0.149}$\\ & 500 & $0.234\pm0.240$ & $0.135\pm0.079$ & $\bm{0.244\pm0.234}$ & $0.239\pm0.236$\\ & 1000 & $0.249\pm0.271$ & $0.271\pm0.256$ & $0.220\pm0.120$ & $\bm{0.294\pm0.243}$\\ & 5000 & $0.276\pm0.253$ & $\bm{0.339\pm0.236}$ & $0.224\pm0.107$ & $0.335\pm0.241$\\ & 10000 & $0.270\pm0.217$ & $0.353\pm0.231$ & $0.237\pm0.105$ & $\bm{0.382\pm0.242}$\\
     \bottomrule
\end{tabular}
\label{tab: f1 on real-world true dags}
\end{table}%
\begin{table}[H]
\centering
\caption{Mean and standard deviation of normalised SHD scores produced by Structural EM, HC-pairwise, HC-IPW and HC-aIPW for real-world networks, under the different assumptions of missingness and sample sizes.}
\begin{tabular}{>{\centering\arraybackslash}m{1cm}>{\centering\arraybackslash}m{1cm}>{\centering\arraybackslash}m{1.9cm}>{\centering\arraybackslash}m{1.9cm}>{\centering\arraybackslash}m{1.9cm}>{\centering\arraybackslash}m{1.9cm}}
    \toprule
     data & sample size & Structural EM & HC-pairwise & HC-IPW & HC-aIPW \\
    \midrule\multirow{5}{*}{MCAR} & 100 & $\bm{1.129\pm0.098}$ & $1.145\pm0.137$ & $1.180\pm0.160$ & $1.151\pm0.136$\\ & 500 & $\bm{0.921\pm0.287}$ & $0.928\pm0.284$ & $0.950\pm0.303$ & $0.927\pm0.284$\\ & 1000 & $1.008\pm0.434$ & $0.883\pm0.303$ & $\bm{0.878\pm0.298}$ & $0.883\pm0.303$\\ & 5000 & $0.975\pm0.437$ & $\bm{0.869\pm0.310}$ & $0.882\pm0.321$ & $\bm{0.869\pm0.310}$\\ & 10000 & $0.909\pm0.368$ & $0.856\pm0.296$ & $\bm{0.854\pm0.291}$ & $0.856\pm0.296$\\\midrule\multirow{5}{*}{MAR} & 100 & $\bm{1.157\pm0.137}$ & $1.161\pm0.150$ & $1.189\pm0.129$ & $1.189\pm0.128$\\ & 500 & $1.096\pm0.217$ & $\bm{0.963\pm0.438}$ & $1.125\pm0.189$ & $0.999\pm0.451$\\ & 1000 & $1.070\pm0.443$ & $\bm{0.937\pm0.472}$ & $0.950\pm0.415$ & $0.942\pm0.411$\\ & 5000 & $1.126\pm0.597$ & $1.027\pm0.453$ & $0.970\pm0.444$ & $\bm{0.937\pm0.436}$\\ & 10000 & $1.224\pm0.551$ & $1.095\pm0.466$ & $0.868\pm0.609$ & $\bm{0.856\pm0.599}$\\\midrule\multirow{5}{*}{MNAR} & 100 & $1.190\pm0.115$ & $1.160\pm0.155$ & $1.122\pm0.229$ & $\bm{1.108\pm0.214}$\\ & 500 & $\bm{1.065\pm0.313}$ & $1.169\pm0.154$ & $1.111\pm0.288$ & $1.097\pm0.277$\\ & 1000 & $1.065\pm0.385$ & $1.064\pm0.356$ & $1.125\pm0.179$ & $\bm{1.045\pm0.344}$\\ & 5000 & $1.139\pm0.424$ & $\bm{1.035\pm0.389}$ & $1.182\pm0.206$ & $1.061\pm0.379$\\ & 10000 & $1.158\pm0.359$ & $1.073\pm0.394$ & $1.222\pm0.218$ & $\bm{1.012\pm0.413}$\\
     \bottomrule
\end{tabular}
\label{tab: shd on real-world true dags}
\end{table}%
\section{Supplementary results of execution time}
\label{app: execution time results}
See results in table~\ref{tab: time}.
\begin{table}[H]
\centering
\caption{Mean and standard deviation of execution times produced by Structural EM, HC-pairwise, HC-IPW and HC-aIPW for sparse networks and relative to HC when applied to complete data, under the different assumptions of missingness and sample sizes.}
\begin{tabular}{>{\centering\arraybackslash}m{1cm}>{\centering\arraybackslash}m{1cm}>{\centering\arraybackslash}m{1.9cm}>{\centering\arraybackslash}m{1.9cm}>{\centering\arraybackslash}m{1.9cm}>{\centering\arraybackslash}m{1.9cm}}
    \toprule
     data & sample size & Structural EM & HC-pairwise & HC-IPW & HC-aIPW \\
     \midrule
     \multirow{5}{*}{MCAR} & 100 & $161.29\pm 54.37$ & $1.55\pm 0.68$ & $8.02\pm 2.27$ & $8.30\pm 1.95$\\
     & 500 & $423.61\pm 146.51$ & $1.55\pm 0.43$ & $6.28\pm 1.50$ & $6.89\pm 1.47$\\
     & 1000 & $556.72\pm 178.34$ & $1.54\pm 0.36$ & $5.88\pm 1.68$ & $6.44\pm 1.78$\\
     & 5000 & $739.52\pm 269.94$ & $1.73\pm 0.70$ & $8.76\pm 2.48$ & $9.24\pm 2.38$\\
     & 10000 & $642.91\pm 254.91$ & $1.63\pm 0.42$ & $11.46\pm 3.45$ & $11.90\pm 3.57$\\
     \midrule
     \multirow{5}{*}{MAR} & 100 & $164.96\pm 46.26$ & $1.64\pm 0.51$ & $8.33\pm 2.15$ & $9.14\pm 2.29$\\
     & 500 & $421.37\pm 144.99$ & $1.62\pm 0.42$ & $7.50\pm 2.00$ & $8.42\pm 2.09$\\
     & 1000 & $554.72\pm 186.89$ & $1.55\pm 0.34$ & $7.73\pm 2.29$ & $8.62\pm 2.47$\\
     & 5000 & $740.71\pm 275.59$ & $1.59\pm 0.32$ & $14.81\pm 5.49$ & $15.33\pm 5.24$\\
     & 10000 & $657.52\pm 313.70$ & $1.73\pm 0.53$ & $22.49\pm 7.52$ & $22.75\pm 7.20$\\
     \midrule
     \multirow{5}{*}{MNAR} & 100 & $154.04\pm 44.86$ & $1.54\pm 0.56$ & $8.80\pm 3.27$ & $8.79\pm 2.67$\\
     & 500 & $419.91\pm 145.50$ & $1.62\pm 0.44$ & $8.18\pm 3.58$ & $7.90\pm 2.47$\\
     & 1000 & $552.16\pm 181.62$ & $1.58\pm 0.41$ & $7.69\pm 3.05$ & $7.56\pm 2.06$\\
     & 5000 & $776.32\pm 359.57$ & $1.87\pm 0.87$ & $13.29\pm 5.93$ & $12.89 \pm 7.74$\\
     & 10000 & $615.91\pm 271.88$ & $1.86\pm 0.77$ & $20.07\pm 9.38$ & $17.93\pm 8.38$\\
     \bottomrule
\end{tabular}
\label{tab: time}
\end{table}
\end{appendices}

\section*{Availability of data and material}
The data used for the simulation results are available upon request to the corresponding author.
\section*{Conflict of interest}
The authors declare that they have no conflict of interest.
\section*{Authors contribution}
All authors contributed to the study conception and design. Material preparation, data collection and analysis were performed by Yang Liu. The first draft of the manuscript was written by Yang Liu and all authors commented on previous versions of the manuscript. All authors read and approved the final manuscript.
\begin{acknowledgements}
This research was supported by the EPSRC Fellowship project EP/S001646/1 on \emph{Bayesian Artificial Intelligence for Decision Making under Uncertainty}.
\end{acknowledgements}

\bibliographystyle{spbasic}      % basic style, author-year citations
\bibliography{manuscript}
\end{document}